\PassOptionsToPackage{numbers,sort&compress}{natbib}
\documentclass{article}
\usepackage[preprint]{corl/corl_2026}
\usepackage{multicol}
\usepackage{xcolor}

\definecolor{urlblue}{HTML}{006EE7}
\definecolor{reflinkred}{HTML}{FE4119}
\definecolor{citegreen}{HTML}{4ADD3B}

\hypersetup{
    pdftitle={FingerEye: Learning Dexterous Manipulation with Continuous Vision-Tactile Sensing},
    pdfauthor={Zhixuan Xu, Yichen Li, Xuanye Wu, Tianyu Qiu, Lin Shao},
    colorlinks=true,
    urlcolor=urlblue,
    linkcolor=urlblue,
    citecolor=urlblue,
    filecolor=urlblue,
}
\usepackage{graphics} 
\usepackage{epsfig} 
\usepackage{mathptmx} 
\usepackage{amsmath} 
\usepackage{amssymb}  
\usepackage{bm}
\usepackage{siunitx} 
\usepackage{textcomp}
\usepackage{gensymb} 
\usepackage{booktabs}
\usepackage{multirow}
\usepackage{colortbl}
\usepackage{xcolor}
\usepackage{algorithm}
\usepackage{algpseudocode}

\usepackage{enumitem}
\usepackage{xspace}
\usepackage{float}
\usepackage{comment}
\usepackage{mathtools}
\usepackage{soul}
\usepackage{adjustbox} 
\usepackage{multicol}
\usepackage{graphicx} 
\usepackage{makecell}
\usepackage{rotating}
\usepackage{wrapfig}
\usepackage{afterpage}
\usepackage[percent]{overpic}

\usepackage{microtype}
\usepackage{contour}
\usepackage{courier}

\usepackage{subcaption} 
\usepackage[font=small]{caption}
\usepackage[percent]{overpic}

\usepackage[11pt]{moresize}

\usepackage{pifont}

\usepackage{xcolor}

\definecolor{MyDarkBlue}{rgb}{0,0.08,1}
\definecolor{airforceblue}{rgb}{0.36, 0.54, 0.66}
\definecolor{MyDarkGreen}{rgb}{0.02,0.6,0.02}
\definecolor{MyDarkRed}{rgb}{0.8,0.02,0.02}
\definecolor{MyDarkOrange}{rgb}{0.40,0.2,0.02}
\definecolor{MyPurple}{RGB}{111,0,255}
\definecolor{MyRed}{rgb}{1.0,0.0,0.0}
\definecolor{MyGold}{rgb}{0.75,0.6,0.12}
\definecolor{MyDarkgray}{rgb}{0.66, 0.66, 0.66}
\definecolor{MyPink}{rgb}{0.9, 0.33, 0.5}
\definecolor{MyCyan}{rgb}{0., 0.4, 0.4}

\definecolor{guidance_green}{RGB}{12,131,27}
\definecolor{theme_orange}{RGB}{255,138,0}
\definecolor{theme_blue}{RGB}{67,99,216}
\definecolor{theme_taro}{RGB}{219,176,234}

\definecolor{pure_green}{RGB}{0,255,0}
\definecolor{pure_red}{RGB}{255,0,0}

\makeatletter
\DeclareRobustCommand\onedot{\futurelet\@let@token\@onedot}
\def\@onedot{\ifx\@let@token.\else.\null\fi\xspace}

\makeatother

\newcommand{\ours}{FingerEye\xspace}

\usepackage[most]{tcolorbox}

\usepackage{xparse}
\usepackage{lipsum}

\DeclareMathAlphabet{\mathcal}{OMS}{cmsy}{m}{n}

\usepackage{etoc}

\def\arxivversion{}
\def\arxivcombined{}
\newcommand{\projectwebsite}{\href{https://nus-lins-lab.github.io/FingerEyeWeb/}{https://nus-lins-lab.github.io/FingerEyeWeb/}}

\begin{document}
\raggedbottom
\title{\LARGE \bf
\ours: Learning Dexterous Manipulation with Continuous Vision-Tactile Sensing
}

\author{
    Zhixuan Xu\textsuperscript{1*},
    Yichen Li\textsuperscript{1,2*},
    Xuanye Wu\textsuperscript{2,3},
    Tianyu Qiu\textsuperscript{2,4},
    Lin Shao\textsuperscript{1,2\textdagger}\\
    \textsuperscript{1}National University of Singapore \quad
    \textsuperscript{2}RoboScience \\
    \textsuperscript{3}Huazhong University of Science and Technology \quad
    \textsuperscript{4}South China University of Technology\\
    \textsuperscript{*}Equal contribution \quad
    \textsuperscript{\textdagger}Corresponding author\\
    \vspace{0.3cm}
    \projectwebsite
}

\maketitle

\begin{abstract}
Dexterous robotic manipulation requires perception that remains informative from pre-contact approach to contact initiation and post-contact control. We introduce FingerEye, a sensing and learning framework that strengthens robotic dexterity through continuous vision-tactile feedback throughout interaction.
On the sensing side, FingerEye integrates binocular RGB cameras with a compliant contact interface to support perception both before and after contact. Before contact, the fingertip cameras provide close-range visual cues and implicit stereo for precise approach and object localization. After contact, marker-tracked deformation of the compliant ring provides a proxy for contact wrench sensing.
On the learning side, we build real-and-sim infrastructure for data collection and evaluation, systematically study policy-interface designs for learning with multiple FingerEye sensors, and develop FingerEye Policy, which applies group-structured modality fusion to reduce modality shortcuts and better exploit distributed fingertip feedback.
Across seven contact-sensitive task settings, FingerEye improves wrist-only policy by over \(30\) percentage points in mean success rate in both simulation and the real world.
\end{abstract}

\section{Introduction}
\begin{figure}[!t]
\centering
\includegraphics[width=\linewidth]{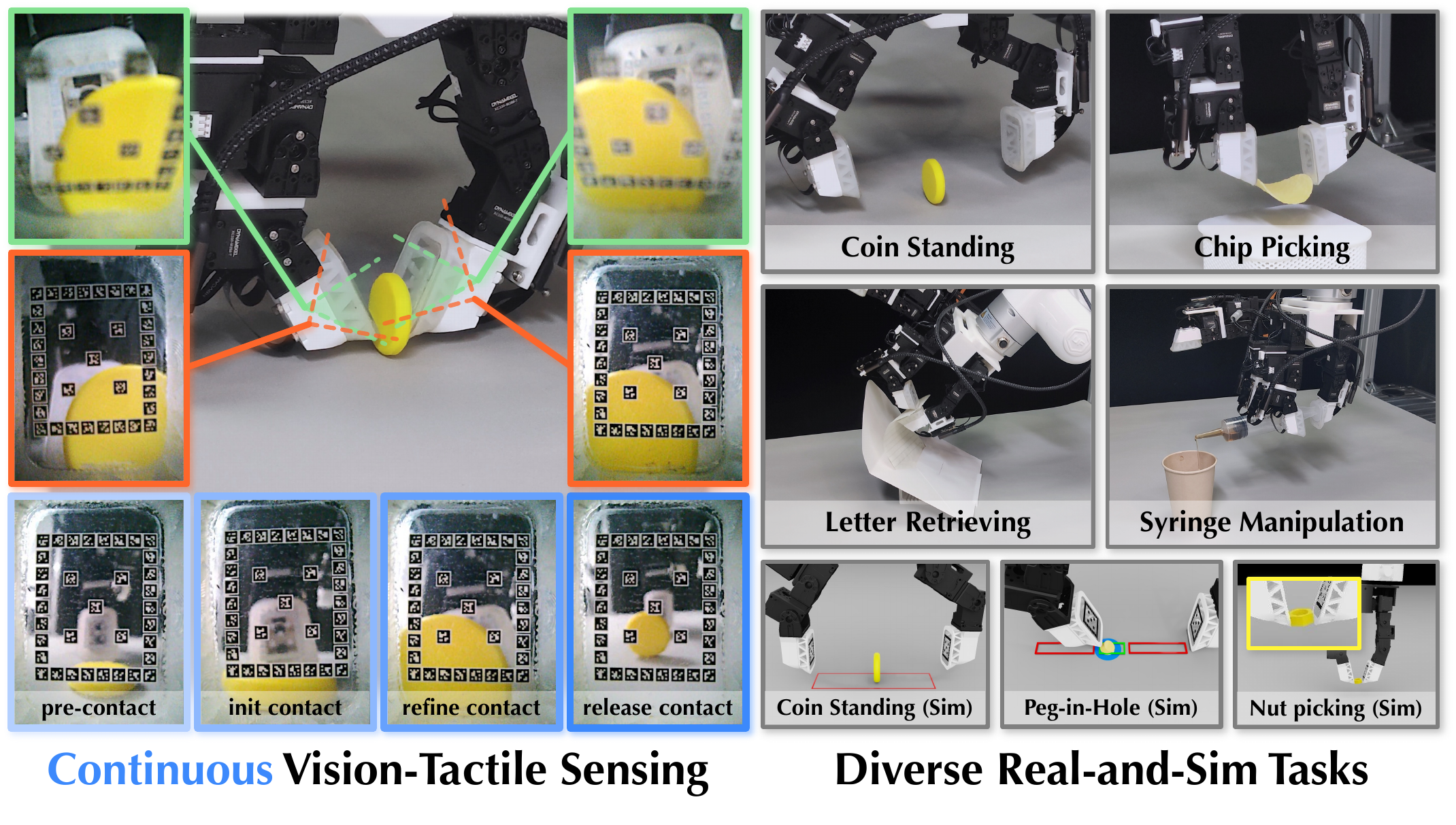}
\vspace{-4mm}
\caption{\textbf{FingerEye overview.}
FingerEye provides continuous fingertip feedback for contact-rich dexterous manipulation. Left: FingerEye tracks local interaction cues throughout approach, contact initiation, contact refinement, and release. Right: we evaluate FingerEye across seven real-world and simulation tasks spanning precise approach, edge/contact engagement, and post-contact stabilization, with objects varying in size, rigidity, delicacy, and geometry.}
\label{fig:task}
\vspace{-3mm}
\end{figure}

``Manipulation refers to an agent’s control of its environment through selective contact.''~\cite{mason2018toward}
In dexterous manipulation, selective contact requires sensing that remains informative before contact, during contact initiation, and after contact.
For example, standing a coin upright on a table (Fig.~\ref{fig:task}) requires the robot to localize the thin coin edge \emph{before contact}, detect fingertip contact onset \emph{during contact initiation}, and regulate contact forces \emph{after contact} to prevent slipping or toppling.
This motivates continuous sensing throughout interaction for precise contact initiation and force regulation.

Despite the importance of continuous sensing throughout interaction, existing systems mainly rely on two separate sources: external vision and tactile sensing. External vision from environment- or wrist-mounted cameras helps robots localize objects, but provides indirect observation to contact~\cite{Ze2024DP3,wang2024dexcap,lin2025sim}. Tactile sensing captures rich contact information, but only after contact is established~\cite{lin2025learning,qi2023general,suresh2024neuralfeels}. 
This separation leaves the most fragile moment in dexterous manipulation—the transition from seeing to touching—weakly observed, limiting precision and dexterity in contact-rich tasks.

Recently, See-Through-Skin (STS) sensors~\cite{yamaguchi2017implementing,athar2023vistac} have emerged as a promising middle ground between vision and tactile sensing. Despite their potential, these approaches are still mostly demonstrated on gripper-centric, low-dexterity manipulation tasks.
What prevents them from reaching the precision and dexterity required by tasks such as coin standing? We identify five subtle yet critical issues: 
1) \textbf{Discontinuous or unnatural visual observations.} Existing designs often suffer from blurry, color-shifted, or discontinuous visual observations due to semi-transparent optical interfaces~\cite{yamaguchi2017implementing,lancaster2022optical,luu2025vision,hogan2022finger}, active or non-natural self-illumination~\cite{wang2022spectac}, illumination switching~\cite{roberge2023stereotac,ablett2025multimodal}, or mechanically reconfigurable components~\cite{xu2025dtactive,dong2026look}. 
2) \textbf{Fragile contact perception in open environments.} Marker- or keypoint-based tactile perception can be unreliable under changing backgrounds, lighting, occlusions, and subtle contact-induced motion~\cite{yamaguchi2016combining,li2026simultaneous}. 
3) \textbf{Limited dexterous-hand compatibility.} Bulky, flat, or front-facing designs can be difficult to mount on compact dexterous fingertips and often miss peripheral or non-frontal contacts in confined spaces. 
4) \textbf{Missing real-and-sim learning infrastructure.} Existing systems often lack integrated data collection and simulation pipelines for learning multisensory policies and evaluating them at scale on dexterous tasks.
5) \textbf{Underdeveloped end-to-end policy interfaces.} Existing STS applications often rely on hand-crafted controllers. For end-to-end learning, naive fusion or conditioning can cause the policy to bypass local \ours feedback and rely on easier but less contact-informative cues.

In this paper, we address these issues through careful design of both the sensing and learning interfaces.
1) We design the right physical \textbf{sensing interface} for dexterous fingertips. \ours combines binocular RGB cameras with a transparent contact surface for natural visual perception and implicit stereo depth before contact. After contact, external forces and torques deform a surrounding compliant ring, whose deformation is captured through robust tag-array pose estimation as a proxy for contact wrench sensing. This design preserves a single RGB sensing pipeline without active illumination or sensing-mode switching, while remaining compact, low-cost, and capable of sensing both frontal and peripheral contacts.
2) We develop a \textbf{learning interface} for multi-sensor \ours feedback. We build real-and-sim infrastructure for data collection, policy training, and scalable evaluation on diverse contact-sensitive tasks, and study how multi-fingertip observations should be fused for action prediction. Since \ours provides local fingertip feedback while wrist vision and proprioception provide easier global cues, naive fusion can cause policies to fit demonstrations while underusing contact-local information. To address this, \ours Policy uses \textbf{group-structured modality fusion}: modality groups are first encoded into structured representations, and action queries then fuse them through group-conditioned decoding to balance cross-modal fusion.

The final system, \textbf{\ours}, provides an integrated sensing and learning framework for contact-rich dexterous manipulation. We evaluate \ours across diverse simulation and real-world tasks, including coin standing, chip picking, letter retrieving, syringe manipulation, nut picking, and peg-in-hole pushing.
\ifdefined\arxivversion
Hardware and software resources are available on the project website: \projectwebsite.
\else
The hardware and software system will be released upon publication.
\fi

\section{Related work}
\textbf{Tactile and See-Through-Skin sensors.}
Tactile sensing has been widely studied for robotic manipulation, including piezoresistive~\cite{wang2019flexible,yu2022allprinted,huang2026flexitac}, piezoelectric~\cite{huang2024high}, biomimetic vibration-sensitive tactile sensors~\cite{fishel2012sensing}, and vision-based tactile sensors~\cite{yuan2017gelsight,lin20239dtact,ward2018tactip,li2024minitac,zhao2025embedding,lin2025lighttact,li2025classification}. Most of these sensors provide rich contact information but are inherently limited to in-contact perception. See-Through-Skin (STS) sensors address this limitation by integrating visual and tactile perception within a single sensing surface~\cite{yamaguchi2017implementing,athar2023vistac,hogan2022finger,roberge2023stereotac,ablett2025multimodal,yamaguchi2016combining,li2026simultaneous}. However, existing STS designs often trade off visual clarity, sensing continuity, contact perception robustness, and fingertip compatibility, due to semi-transparent interfaces~\cite{luu2025vision,hogan2022finger,luu2023soft}, active illumination or visual-tactile mode switching~\cite{wang2022spectac,roberge2023stereotac,ablett2025multimodal}, mechanically reconfigurable components~\cite{dong2026look}, or marker/keyline tracking under open-background visual noise~\cite{yamaguchi2016combining,li2026simultaneous}. In contrast, \ours targets continuous dexterous fingertip sensing, combining natural close-range binocular RGB perception before contact with robust deformation sensing after contact.

\textbf{Multisensory policy learning.}
Recent work has explored robot policies that combine multi-view vision, touch, force/torque, audio, and proprioception for contact-rich manipulation~\cite{li2022see,liu2024maniwav,feng2024play}. A line of work improves tactile or visuo-tactile representations through self-supervised or predictive pretraining~\cite{higuera2024sparsh,higuera2025tactilebeyondpixels,heng2025vitacformer,liu2025vitamin,zhu2026touch}. At the policy-interface level, many methods adopt flat multimodal fusion: sensory tokens are concatenated and processed by a shared encoder~\cite{zhao2023act,heng2025vitacformer,li2022see,zhao2025polytouch}, or action tokens attend to a mixed pool of encoded sensory context in diffusion- or chunking-based policies~\cite{lin2025learning,chi2025diffusion,xu2025robopanoptes,jiang2025gelfusion}. While effective in many settings, direct fusion can underuse sparse but task-critical contact signals when global visual or proprioceptive cues provide easier shortcuts~\cite{chen2025multi,liu2025factr,lei2026learning}. In contrast, we study policy interfaces for distributed \ours feedback and introduce \emph{group-structured modality fusion}: modality groups are first encoded into structured representations, and action queries then fuse them through group-conditioned decoding to reduce cross-modal attention competition.

\begin{figure}[!t]
\begin{minipage}[b]{0.42\linewidth}
\centering
\includegraphics[width=1.0\linewidth]{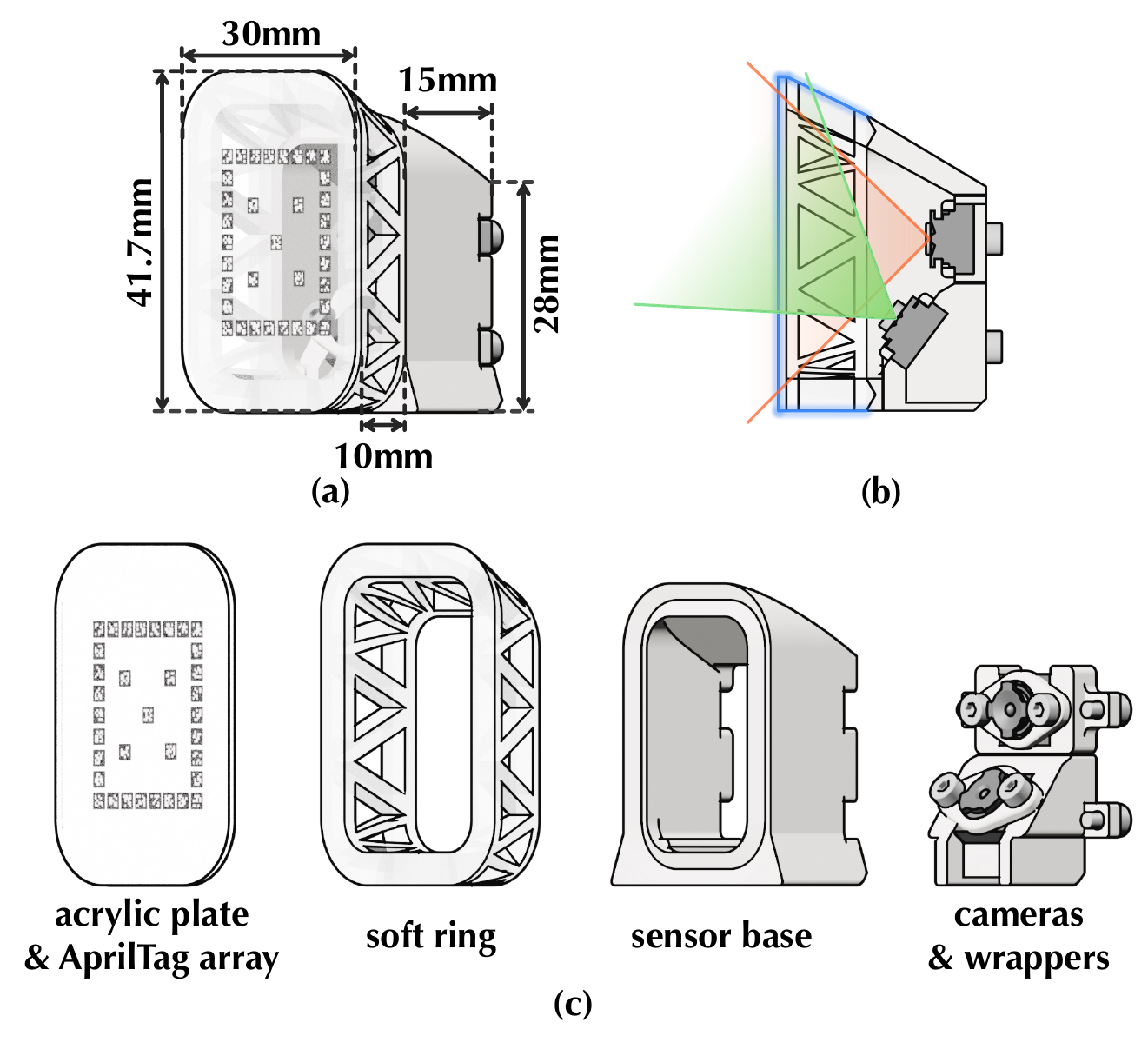}
\caption{
\textbf{\ours Sensing Interface Design.}
(a) Dimensions. 
(b) Cross-sectional view: fields of view of the \textcolor[RGB]{131,226,145}{root} and \textcolor[RGB]{253,102,43}{tip} cameras, and the \textcolor[RGB]{50,131,254}{contact sensing region}. 
(c) Exploded view.
}
\label{fig:design}
\end{minipage}
\hfill
\begin{minipage}[b]{0.54\linewidth}
\centering
\begin{minipage}[t]{0.88\linewidth}
\centering
\includegraphics[width=\linewidth]{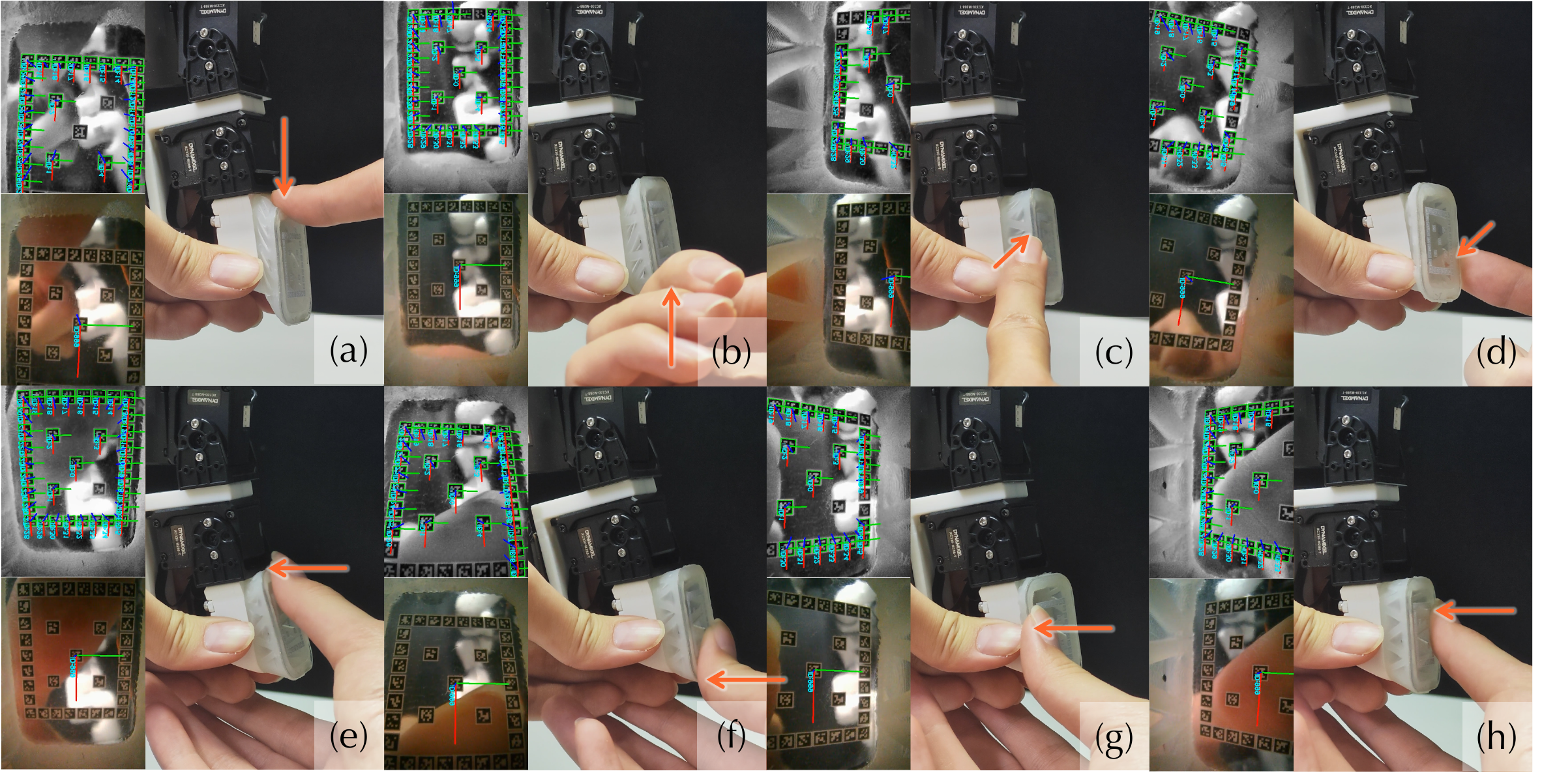}
\includegraphics[width=\linewidth]{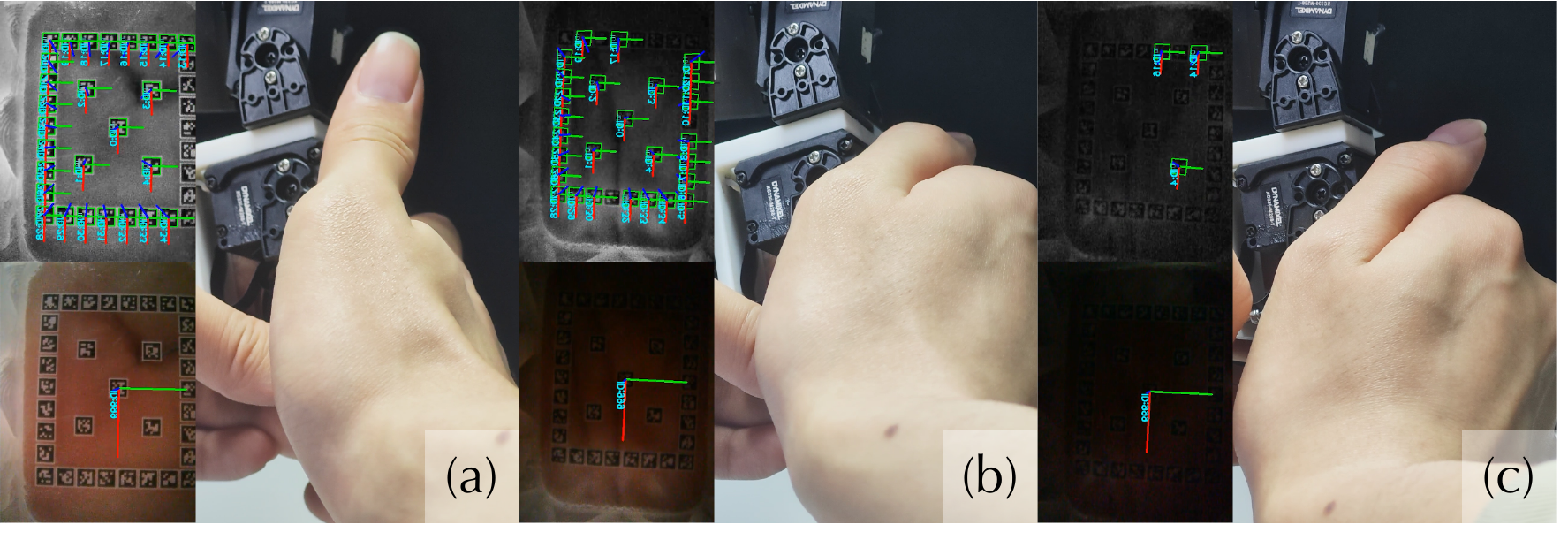}
\end{minipage}
\caption{\textbf{Robustness of Tag Array Pose Estimation.} Top: stable detection under frontal and peripheral force perturbations. Bottom: stable detection under lighting variation.}
\label{fig:robustness_main}
\end{minipage}
\vspace{3mm}
\end{figure}

\section{\ours Sensing Interface}
\label{sec:sensing}
The key sensing-interface question is: how can a fingertip sensor see clearly and continuously before contact, sense contact robustly after contact, and still fit naturally on a dexterous hand? We address this through five design highlights and four capability highlights.

\subsection{Sensing Interface Design Highlights}
\label{subsec:sensing_design}

\textbf{DH1. Complementary binocular RGB cameras for wide view and implicit depth.}
Each \ours module integrates two RGB cameras with complementary viewing ranges and roles (Fig.~\ref{fig:design}). The tip-facing camera observes the transparent contact surface at a short working distance of approximately \SI{10}{mm}, enabling tag-array pose estimation after contact, while the root-side camera operates at a longer working distance and is tilted to preserve a broader view during approach. Together, they provide a larger effective field of view and implicit stereo depth cues for object localization and contact alignment, while preserving natural RGB observations without active illumination or sensing-mode switching. This complementary binocular design is important for policy learning: our experiments show that binocular \ours observations improve manipulation success rates.

\textbf{DH2. Peripheral compliant ring for contact sensing without blocking vision.}
Instead of placing a deformable gel directly in the visual path, \ours keeps a clear central acrylic window and surrounds it with a compliant silicone ring. External forces and torques deform the ring and induce measurable plate-pose changes while leaving the central RGB view unobstructed. This peripheral compliant interface lets both frontal and peripheral contacts affect the estimated plate pose, expanding the effective sensing region for pushing, wedging, and delicate grasping.

\textbf{DH3. Multi-tag pose tracking for robust deformation sensing.}
\label{sec:def_est}
A custom AprilTag array on the acrylic plate converts contact-driven ring deformation into a compact six-dimensional plate-pose change. We aggregate visible tag corners into one multi-tag PnP problem, instead of combining single-tag pose estimates. When some tags are invisible under force or lighting perturbations, the remaining corners provide redundant constraints for stable pose tracking.
\ifdefined\arxivversion
Estimator details and comparison with keyline marker tracking~\cite{li2026simultaneous} are provided in Sec.~\ref{subsec:appendix_marker} and Sec.~\ref{subsec:supp_gelsight}.
\else
Details and comparison with keyline marker tracking~\cite{li2026simultaneous} are provided in the Supp.
\fi

\textbf{DH4. Compact wedge-shaped form factor for dexterous fingertips.}
The assembled module adopts a wedge-shaped, fingertip-scale form factor with an overall size of approximately \(41.7\times30.0\times25.0\) \SI{}{mm}, making \ours compatible with compact dexterous hands and confined manipulation spaces.

\textbf{DH5. Low-cost and reproducible implementation.}
\ours uses only low-cost off-the-shelf and 3D-printed components, with a bill of materials of roughly \$60 per module;
\ifdefined\arxivversion
fabrication details are provided in Sec.~\ref{subsec:supp_fabrication}, and hardware/software resources are available on the project website: \projectwebsite.
\else
fabrication details are provided in the Supp. and will be released upon publication.
\fi

\begin{figure}[!t]
\centering
\includegraphics[width=1.0\linewidth]{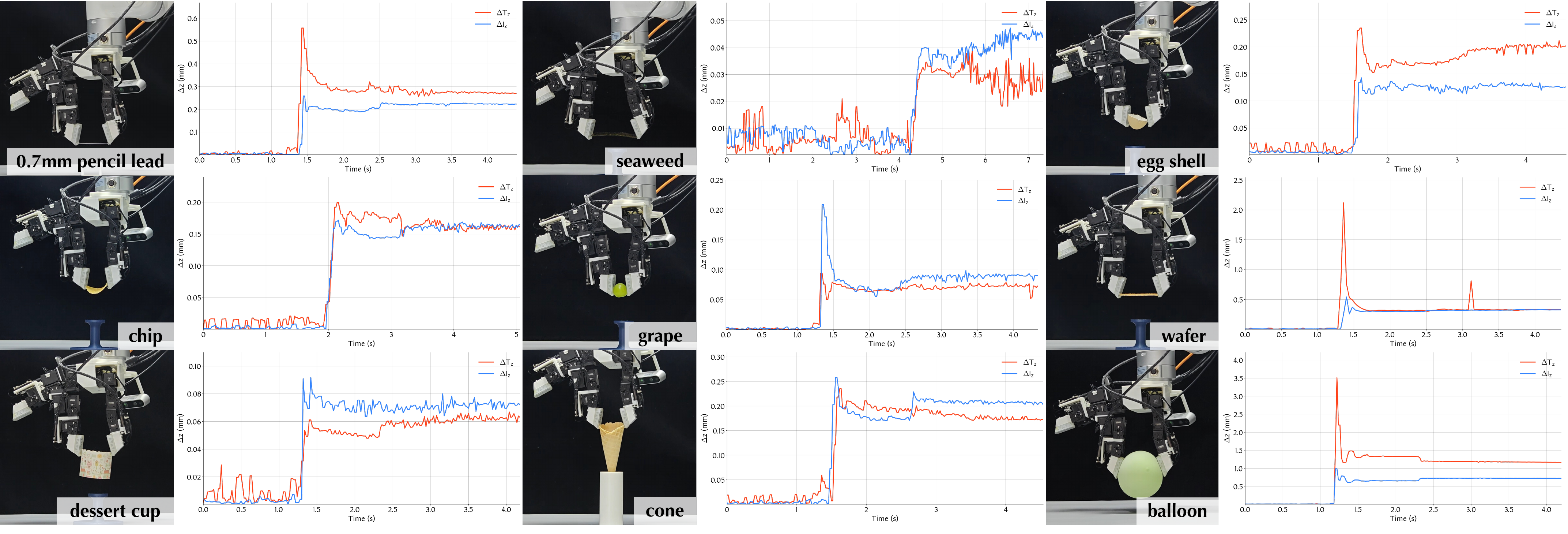}
\caption{
\textbf{Visualization of delicate grasping and fingertip normal deformation.} Each example shows the lifted object and temporal curves of normal displacement $\Delta z$ measured at the \textcolor[RGB]{255,66,24}{thumb} and \textcolor[RGB]{50,131,254}{index} fingertips.
}
\label{fig:delicate_grasp_exp}
\vspace{1mm}
\end{figure}

\subsection{Sensing Interface Capability Highlights}
\label{subsec:sensing_capabilities}

\textbf{CH1. Robust pose estimation under force and lighting variations.}
\label{subsec:robustness}
As shown in Fig.~\ref{fig:robustness_main}, even when force or lighting variation makes some tags invisible, the remaining visible corners still provide redundant constraints for stable multi-tag PnP, supporting the tag-array design in \textbf{DH3}. A tag-count ablation quantifies the tracking--coverage trade-off (Sec.~\ref{subsec:supp_tag_count}). After \(10{,}000\) normal-compression cycles (\(3\) mm, \(1\) Hz), we observe no cracking or delamination, with largely retained task performance (Sec.~\ref{subsec:supp_durability}).

\textbf{CH2. Plate-pose change as a contact-wrench proxy.}
\label{subsec:relation}
Across synchronized wrench--pose pairs, the estimated six-dimensional plate-pose change reliably predicts the applied wrench. The high held-out \(R^2_{\rm test}\) validates it as a proxy for contact wrench sensing;
\ifdefined\arxivversion
details are provided in Sec.~\ref{subsec:supp_corr_setup}.
\else
details are provided in the Supp.
\fi

\textbf{CH3. Analytical sensitivity to small contact wrenches.}
\label{subsect:sensitivity}
Following detection sensitivity analysis~\cite{ouyang2020low,zhu2025shapeforce}, \ours achieves minimum detectable force \(\mathbf{F}_{\min}=[\,4.30,\,4.22,\,9.93\,]~\mathrm{mN}\) and torque \(\boldsymbol{\tau}_{\min}=[\,0.32,\,0.13,\,8.55\,]~\mathrm{mN\cdot m}\).
\ifdefined\arxivversion
Derivations are provided in Sec.~\ref{subsec:supp_sensitivity}.
\else
Derivations are provided in the Supp.
\fi

\textbf{CH4. Contact-onset utility in delicate grasping.}
\label{subsect:delicate}
A Leap Hand~\cite{shaw2023leap} equipped with \ours stops each finger when fingertip normal deformation \((\Delta z)\) indicates contact onset. Across nine fragile or deformable objects, \ours enables contact-aware stopping and damage-free lifting. Fig.~\ref{fig:delicate_grasp_exp} illustrates plate displacement down to \(\Delta z=0.04\) mm; stopping thresholds are selected once per object based on fragility and weight (Sec.~\ref{subsec:supp_delicate_setup}). Unified contact detection remains future work.

\section{\ours Learning Interface}
\label{sec:learning}

We now study how to design and validate a learning interface that improves dexterous manipulation with \ours observations. Our interface combines a real-and-sim infrastructure for collecting contact-sensitive task demonstrations and scalable evaluation with a structured policy design that better leverages local \ours feedback and supports efficient multi-camera training.

\subsection{Real-and-Sim Learning Infrastructure}
\label{subsec:learning_infrastructure}

\textbf{Robot platform.}
Our real-world setup uses a fixed-base uFactory xArm7 equipped with a LEAP Hand~\cite{shaw2023leap}. We provide complementary observations with a wrist-mounted RGB camera for global scene context and \ours modules on each fingertip for local feedback.

\begin{wrapfigure}[14]{r}{0.47\columnwidth}
\centering
\vspace{-7mm}
\includegraphics[width=\linewidth]{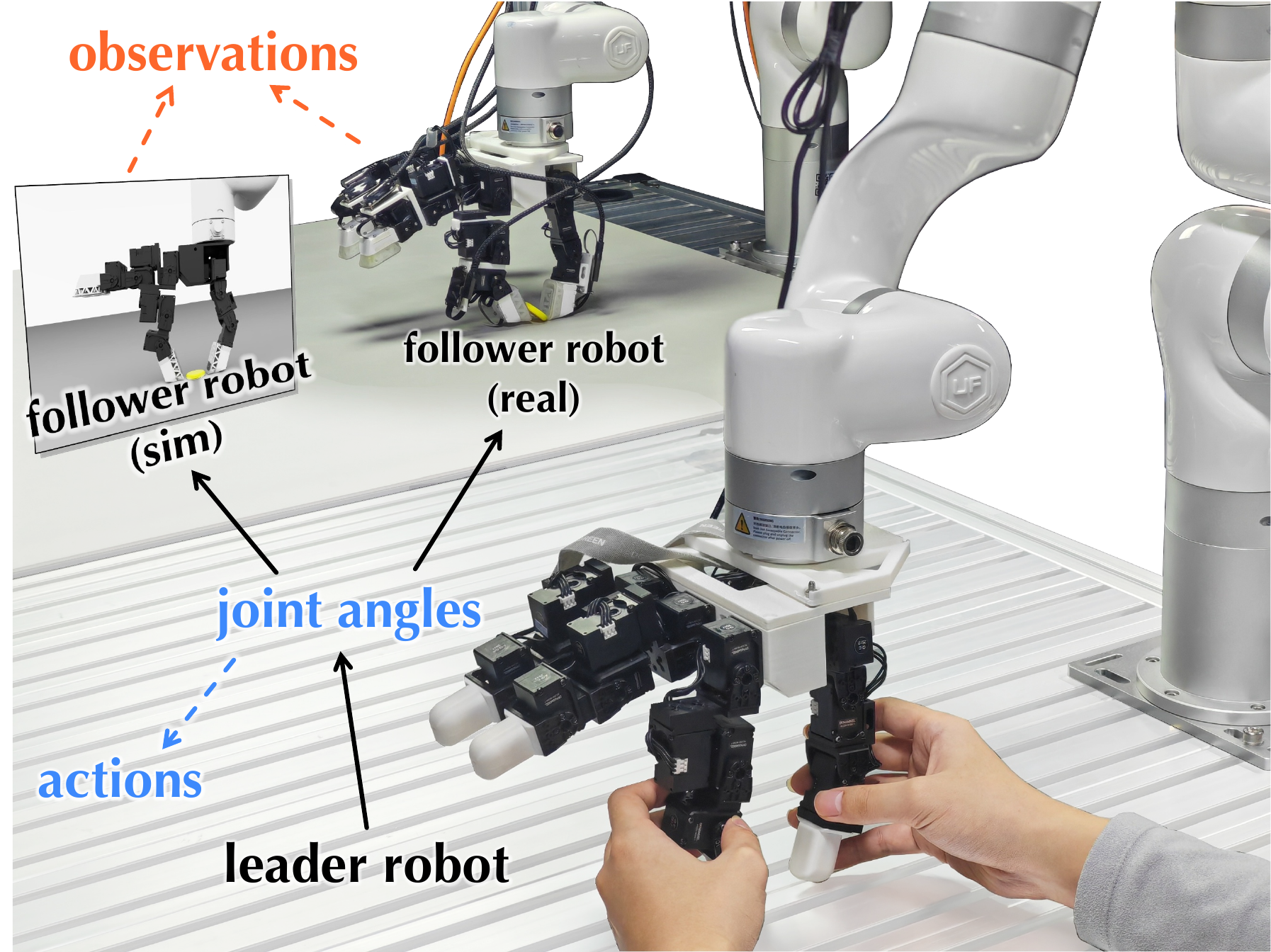}
\caption{\textbf{Data collection interface.}
A leader--follower setup enables intuitive and precise data collection of contact-sensitive tasks in both real and sim.
}
\label{fig:data_collection_interface}
\end{wrapfigure}
\textbf{Data collection interface.}
We collect demonstrations with a leader--follower teleoperation setup using matched hand kinematics. Although retargeting-based alternatives are convenient~\cite{qin2023anyteleop}, we found in our preliminary trials that the embodiment gap between human and robot hands was difficult to tolerate in contact-sensitive tasks. We therefore use a passive matched leader hand, while the follower tracks streamed joint positions through PD control. The system records synchronized wrist images, multi-fingertip \ours observations, plate poses, robot states, and actions at 10\,Hz.

\textbf{Simulation digital twin.}
For scalable evaluation and ablation, we construct a simulation counterpart of the real sensorized hand. Each \ours acrylic plate is modeled as a floating body connected to the sensor base by six virtual joints. The joint gains are calibrated from the measured real-world pose-to-wrench relation, to approximate the plate contact response. The simulator matches geometry, kinematics, camera parameters, and tag layout, but does not model or validate nonlinear silicone mechanics or deformable interactions. This simulation interface enables parallelized policy evaluation and controlled comparisons of sensing and policy-interface designs under reproducible settings.

\subsection{Policy Interface Design Highlights}
\label{subsec:policy_design}
\textbf{Motivation.}
A key challenge in learning dexterous manipulation with \ours is not merely adding more sensory streams, but designing a policy interface that lets local fingertip feedback actively influence the decoded actions. In policy learning, wrist-view observations and proprioception provide strong global cues and can dominate the learning objective, making multimodal policies prone to shortcut reliance. As a result, contact-local cues from \ours, such as contact alignment, edge engagement, and contact-state transitions, may be underused during fusion and decoding. We address this with \textbf{group-structured modality fusion}, which keeps distributed fingertip observations explicit through both encoding and decoder conditioning.

\textbf{Notation.}
Formally, let $\bm{O}_t=(\bm{o}_t^1,\ldots,\bm{o}_t^M)$ denote the observation tuple at time $t$, with $M$ observation groups including wrist RGB, robot proprioception, per-fingertip \ours RGB. The main learned policies use neither plate pose nor estimated wrench; plate pose is used for sensor characterization and delicate-grasp control. We separately evaluate an auxiliary wrench group in Sec.~\ref{subsec:supp_wrench_input}. The policy predicts an action chunk $\bm{A}_t=\pi_\theta(\bm{O}_t)=(\bm{a}_t,\ldots,\bm{a}_{t+T_p-1})$ and executes the first $T_a$ actions before replanning~\cite{zhao2023act,chi2025diffusion}. At \(10\) Hz, most tasks predict \(1.6\)-s chunks and replan every \(0.8\) s; task-specific settings are in Sec.~\ref{subsec:supp_task_settings}.
\ifdefined\arxivcombined
Full tensor-level architecture details, including tokenization and ablation notation, are provided in Sec.~\ref{sec:supp_policy_arch}.
\else
Full tensor-level architecture details, including tokenization and ablation notation, are provided in the supplementary material.
\fi

\textbf{PD1. Modality-group encoding for structured representations.}
Each observation group $\bm{o}_t^i$ is tokenized as $\bm{U}_t^i=\tau_i(\bm{o}_t^i)$, with view-identity embeddings and temporal-identity embeddings; let $\bm{U}_t=[\bm{U}_t^1,\ldots,\bm{U}_t^M]$ denote all policy tokens. We compare three encoder interfaces:
$$
\begin{array}{@{}r@{\;}c@{\quad}r@{\;}c@{\;}l@{}}
\text{\normalfont\bfseries\ttfamily NoEnc}
&:&
\bm{Z}_t^{\mathrm{noenc}}
&=&
\bm{U}_t, \\
\text{\normalfont\bfseries\ttfamily FlatEnc}
&:&
\bm{Z}_t^{\mathrm{flat}}
&=&
E_{\mathrm{flat}}(\bm{U}_t), \\
\text{\normalfont\bfseries\ttfamily GroupEnc}
&:&
\bm{Z}_t^{\mathrm{group}}
&=&
[E_1(\bm{U}_t^1),\ldots,E_M(\bm{U}_t^M)].
\end{array}
$$
\textbf{\texttt{NoEnc}}~\cite{chi2024universal,xu2025robopanoptes,lin2025learning,li2026simultaneous} provides no pre-decoder interaction among observation tokens, while \textbf{\texttt{FlatEnc}}~\cite{zhao2023act,heng2025vitacformer} immediately mixes all local and global cues in a shared encoder. In contrast, \textbf{\texttt{GroupEnc}} first performs within-group interaction, allowing each observation group to form a dedicated representation before cross-group fusion. For \ours, this lets fingertip-local views integrate local geometry and contact-state cues before being fused with wrist vision and proprioception.

\textbf{PD2. Group-conditioned decoding for balanced modality fusion.}
Even after group encoding, local fingertip information can still be underused during decoding: a standard Transformer decoder conditions each action query on a single sensory context, where all modality groups compete in one cross-attention operation. Formally, the decoder starts from action-query tokens $\bm{q}=(\bm{q}^1,\ldots,\bm{q}^{T_p})$ and updates their hidden states $\bm{h}_t^l$ across layers before projecting them to the action chunk $\bm{A}_t$. Let $\bm{Z}_t=[\bm{z}_t^1,\ldots,\bm{z}_t^M]$ denote the encoded sensory context with $M$ modality groups. The flat and group-conditioned decoders differ in the cross-attention update at layer $l$:
\[
\begin{array}
{@{}r@{\;}c@{\quad}r@{\;}c@{\;}l@{}}
\text{\normalfont\bfseries\ttfamily FlatDec CrossAttn}
&:&
\tilde{\bm{h}}_t^l
&=&
\bm{h}_t^l+\mathrm{CrossAttn}^l(\bm{h}_t^l,\bm{Z}_t), \\
\text{\normalfont\bfseries\ttfamily GroupDec CrossAttn}
&:&
\tilde{\bm{h}}_t^l
&=&
\bm{h}_t^l+\frac{1}{M}\sum_{i=1}^{M}
\mathrm{CrossAttn}_i^l(\bm{h}_t^l,\bm{z}_t^i).
\end{array}
\]
In \textbf{\texttt{FlatDec}}~\cite{chi2024universal,xu2025robopanoptes,li2026simultaneous,zhao2023act,heng2025vitacformer}, all modality tokens share one cross-attention normalization per action query, which can bias decoder updates toward easier global cues. In \textbf{\texttt{GroupDec}}, each modality group produces a separate residual update, and we average these residuals before adding them to the action-query states. This keeps action prediction jointly conditioned on all groups while giving each group a direct path into the decoder. For \ours, this reduces competition between fingertip-local observations and global wrist or proprioceptive cues.

\textbf{PD3. Cached visual summaries for efficient multiview training.}
To make multi-camera policy training efficient, we freeze the RADIO image-backbone~\cite{ranzinger2024radio} and cache per-view visual summaries offline. Policy training then uses these cached summaries as visual tokens, avoiding repeated image-backbone inference and enabling efficient training and ablations.

\section{Experiments}
\label{sec:experiments}

\begin{figure*}[t]
\centering
\includegraphics[width=0.98\linewidth]{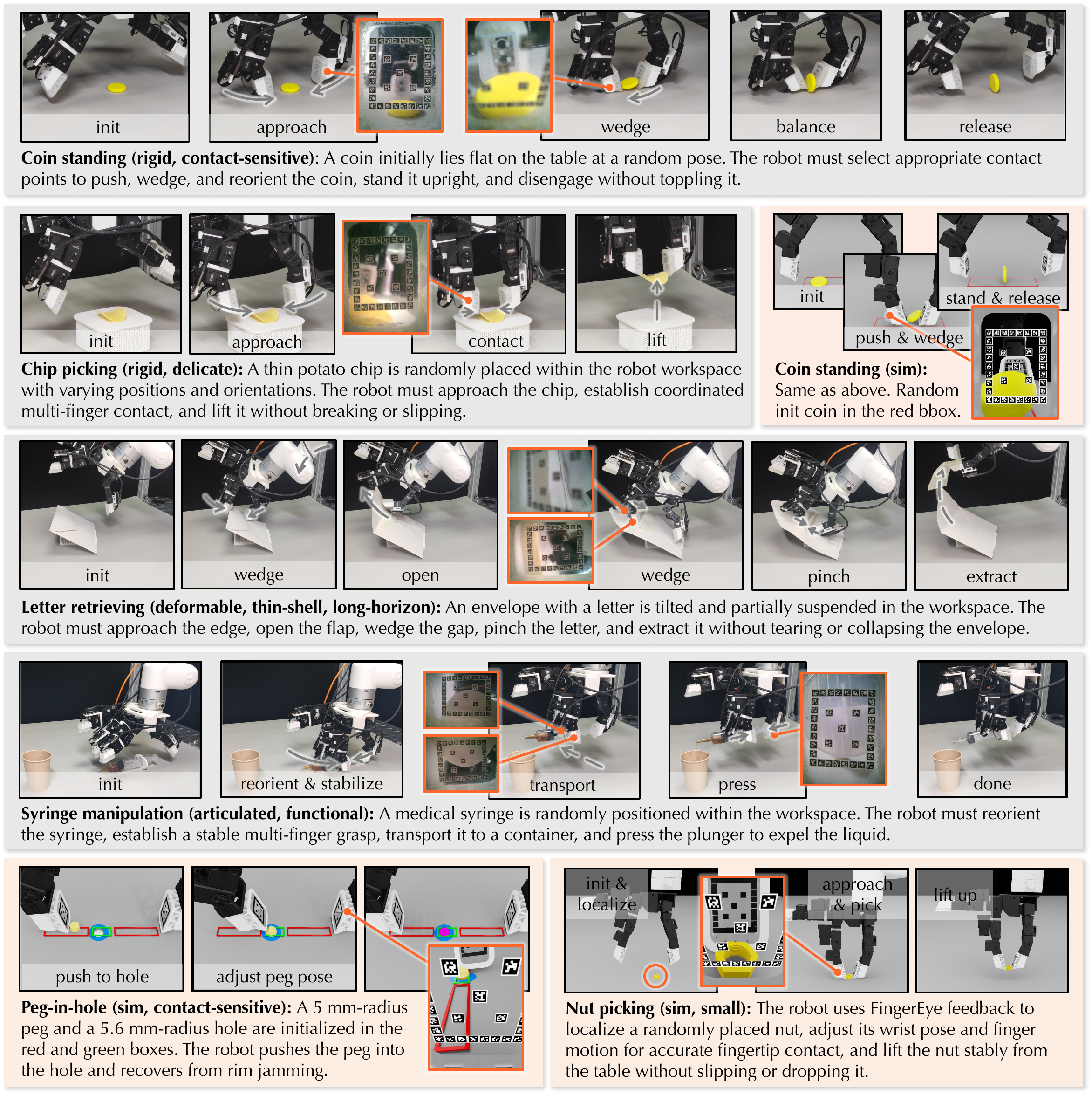}
\caption{
\textbf{\ours task overview.}
We evaluate \ours across seven \textcolor{gray}{real-world} and \textcolor{orange}{simulation} tasks spanning precise approach, edge or contact engagement, and post-contact reorientation or stabilization, with objects that vary in size, rigidity, delicacy, and geometry. The red bounding box marks the coin initialization region.
}
\label{fig:rollout_vis}
\ifdefined\arxivversion
\else
\vspace{-1mm}
\fi

{\captionsetup{type=table,hypcap=false}
\small
\setlength{\tabcolsep}{3pt}
\renewcommand{\arraystretch}{1.0}
\caption{
\textbf{Modality comparison across simulation and real-world tasks.}
\textbf{\texttt{W}} = wrist camera, \textbf{\texttt{M}} = tip-only monocular \ours camera, and \textbf{\texttt{FE}} = binocular \ours cameras; proprioception is included by default.
SR in \%; simulation uses mean$\pm$std and real-world tasks use \(20\) trials.
Green values are percentage-point gains over \textbf{\texttt{W}}.
}
\label{tab:sim_modality}
\centering
\begin{tabular}{@{}lccccccccc@{}}
\toprule
\multirow{2}{*}{Modalities}
& \multicolumn{4}{c}{Simulation}
& \multicolumn{5}{c}{Real-world} \\
\cmidrule(lr){2-5}\cmidrule(lr){6-10}
& Coin & Nut Pick & Peg-in-Hole & Mean & Coin & Chip & Letter & Syringe & Mean \\
\midrule
\textbf{\texttt{W}} & \(25.5{\pm}8.3\) & \(38.7{\pm}23.2\) & \(16.0{\pm}6.6\) & \(26.7\) & \(25\) & \(35\) & \(30\) & \(60\) & \(37.5\) \\
\textbf{\texttt{W+M}} & \(56.4{\pm}3.7\) & \(84.4{\pm}5.2\) & \(36.5{\pm}5.5\) & \(59.1\) {\textcolor{green!50!black}{\scriptsize(+32.4)}} & \(50\) & \(55\) & \(45\) & \(75\) & \(56.3\) {\textcolor{green!50!black}{\scriptsize(+18.8)}} \\
\textbf{\texttt{W+FE}} & \(\mathbf{65.6{\pm}4.8}\) & \(\mathbf{88.2{\pm}4.2}\) & \(\mathbf{43.8{\pm}4.2}\) & \(\mathbf{65.9}\) {\textcolor{green!50!black}{\scriptsize(+39.1)}} & \(\mathbf{65}\) & \(\mathbf{70}\) & \(\mathbf{65}\) & \(\mathbf{85}\) & \(\mathbf{71.3}\) {\textcolor{green!50!black}{\scriptsize(+33.8)}} \\
\bottomrule
\end{tabular}
}
\ifdefined\arxivversion
\else
\vspace{-1mm}
\fi
\end{figure*}

We evaluate whether \ours improves dexterous manipulation by providing continuous vision-tactile observations that complement external vision.
The experiments answer four questions:
\textbf{Q1: \ours benefits.} Does incorporating \ours improve policy performance over external-vision-only observations?
\textbf{Q2: Binocular sensing.} Does the binocular camera design of \ours provide measurable advantages over a monocular configuration?
\textbf{Q3: Continuous sensing.} Does continuous vision-tactile sensing improve over post-contact-only signals?
\textbf{Q4: Policy interface ablations.} Which policy interface best exploits distributed \ours feedback?

\textbf{Tasks and Evaluation Protocol.}
\label{subsec:exp_setup}
Fig.~\ref{fig:rollout_vis} summarizes the simulation and real-world tasks.
Simulation uses \(3\) training seeds and \(3\) evaluation seeds with \(64\) rollouts per evaluation seed.
Real-world evaluation uses \(20\) rollouts per method and task.
We report task success rate (SR). 
For compact notation, we abbreviate the policy variants defined in Sec.~\ref{subsec:policy_design} as \textbf{\texttt{FEnc}},
\textbf{\texttt{GEnc}}, \textbf{\texttt{FDec}}, and \textbf{\texttt{GDec}}, respectively. We size-match all variants to keep
parameter counts comparable.

\subsection{Q1--Q2: Sensing Modality Comparison}
\label{subsec:q1q2_modality}

We first compare observation modalities while keeping the policy interface fixed.
All policy inputs include robot proprioception by default.
\textbf{\texttt{W}} denotes wrist camera;
\textbf{\texttt{M}} denotes the tip-camera-only monocular \ours variant;
and \textbf{\texttt{FE}} denotes full binocular \ours cameras.
For these modality ablations, all methods use the same \textbf{\texttt{GEnc+GDec}} policy architecture.

\begin{figure*}[t]
\centering
\includegraphics[width=\linewidth]{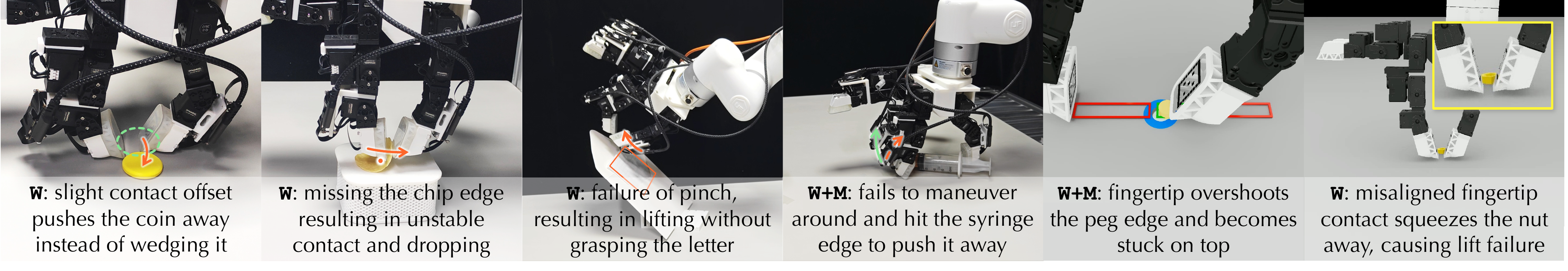}
\caption{
\textbf{Modality failure modes.}
The panels summarize local contact-geometry errors for \textbf{\texttt{W}} and residual monocular coverage errors for \textbf{\texttt{W+M}}; full binocular \ours reduces these failures with complementary views.
}
\label{fig:failure_cases}
\end{figure*}

\textbf{O1: FingerEye visual feedback improves dexterous manipulation beyond wrist-only observation.}
As shown in Table~\ref{tab:sim_modality}, adding binocular \ours to wrist observations improves mean SR over \textbf{\texttt{W}} by \(39.1\) and \(33.8\) percentage points in simulation and real-world tasks, respectively.
The gain is largest where precise fingertip placement matters: in nut picking, the wrist-only policy is about \(50\) percentage points lower in SR and has high variance (\(38.7{\pm}23.2\%\)), indicating that external vision alone struggles with occlusion, final wrist refinement, and coordinated finger motion around small objects.
Fig.~\ref{fig:failure_cases} shows the same pattern qualitatively: \textbf{\texttt{W}} failures come from missing local contact geometry, such as offset coin wedging, missed chip/letter edges, and nut squeeze-out.

\textbf{O2: Binocular FingerEye feedback reduces monocular ambiguity near contact.}
Table~\ref{tab:sim_modality} shows that \textbf{\texttt{W+FE}} improves over \textbf{\texttt{W+M}} by \(6.8\) and \(15.0\) percentage points in simulation and real-world mean SR, respectively.
The remaining \textbf{\texttt{W+M}} failures in Fig.~\ref{fig:failure_cases} show why one fingertip view is still limited: the policy pushes against the syringe edge instead of maneuvering around it, and the fingertip overshoots the peg edge and becomes stuck on top.
These errors are consistent with limited local coverage and single-view ambiguity near contact; complementary fingertip views help resolve edge side, contact alignment, and object engagement.
The same ordering, \textbf{\texttt{W+FE}} \(>\) \textbf{\texttt{W+M}} \(>\) \textbf{\texttt{W}}, appears in both simulation and real-world tasks, strengthening the modality conclusion for Q1--Q2.

\subsection{Q3: Continuous Sensing vs. Post-Contact Signals}
\label{subsec:q3_continuous_sensing}

\begin{table}[H]
\centering
\caption{
\textbf{Continuous vs. post-contact sensing.}
All methods use \textbf{\texttt{GEnc+GDec}} and proprioception.
SR in \%; simulation reports mean$\pm$std, and real-world tasks use \(20\) trials per method and task. \textbf{\texttt{T}} denotes simulated contact-depth maps in simulation and GelSight observations in the real world.
}
\label{tab:post_contact}
\small
\setlength{\tabcolsep}{5pt}
\begin{tabular}{@{}lcccccc@{}}
\toprule
\multirow{2}{*}{Modalities} & \multicolumn{4}{c}{Simulation} & \multicolumn{2}{c}{Real-world} \\
\cmidrule(lr){2-5}\cmidrule(lr){6-7}
& Coin & Nut Pick & Peg-in-Hole & Mean & Chip & Syringe \\
\midrule
\textbf{\texttt{W}} & \(25.5{\pm}8.3\) & \(38.7{\pm}23.2\) & \(16.0{\pm}6.6\) & \(26.7\) & \(35\) & \(60\) \\
\textbf{\texttt{W+T}} & \(27.3{\pm}6.8\) & \(28.6{\pm}5.1\) & \(16.5{\pm}7.4\) & \(24.1\) & \(45\) & \(65\) \\
\textbf{\texttt{W+FE}} & \(\mathbf{65.6{\pm}4.8}\) & \(\mathbf{88.2{\pm}4.2}\) & \(\mathbf{43.8{\pm}4.2}\) & \(\mathbf{65.9}\) & \(\mathbf{70}\) & \(\mathbf{85}\) \\
\bottomrule
\end{tabular}
\end{table}

We compare \textbf{\texttt{W+FE}} against post-contact tactile input \textbf{\texttt{W+T}} in simulation and the real world, using the same \textbf{\texttt{GEnc+GDec}} policy interface (Table~\ref{tab:post_contact}).

\textbf{O3: Continuous fingertip vision is more effective than post-contact-only tactile feedback.}
In simulation, \textbf{\texttt{W+T}} remains close to \textbf{\texttt{W}} in mean SR (\(24.1\%\) vs. \(26.7\%\); Table~\ref{tab:post_contact}), while \textbf{\texttt{W+FE}} reaches \(65.9\%\).
This gap is consistent with the task structure: Coin standing and Peg-in-Hole require precise fingertip pose refinement before stable contact, while Nut Pick requires selecting pinch locations and maintaining squeeze direction on a small object.
These maps encode nearby collision geometry rather than contact force or deformable physics (Sec.~\ref{subsec:supp_post_contact_tactile}) and are silent before near contact, whereas \ours supplies local visual feedback during approach, contact initiation, and contact maintenance, helping the policy decide where and how to engage the object instead of only reacting after contact.

\textbf{Real-world comparison with GelSight.}
With matched demonstrations and policy architecture, \textbf{\texttt{W+T}} (wrist RGB plus GelSight) achieves \(45\%\) SR on Chip Picking and \(65\%\) on Syringe Manipulation, compared with \(70\%\) and \(85\%\) for \textbf{\texttt{W+FE}} (Table~\ref{tab:post_contact}). GelSight's bulk and non-wedge-shaped tip prevent valid demonstrations for Coin Standing and Letter Retrieval; we therefore compare only the feasible tasks. This comparison reflects both sensing and fingertip geometry; a matched STS policy comparison remains a limitation.

\subsection{Q4: Policy Interface for Distributed \ours Feedback}
\label{subsec:q4_policy_interface}

\begin{wraptable}[8]{r}{0.5\columnwidth}
\centering
\vspace{-5mm}
\caption{
\textbf{Policy-interface comparison.}
\textbf{\texttt{W+FE}} input; simulation SR in \%, mean$\pm$std.
}
\label{tab:sim_arch}
\scriptsize
\setlength{\tabcolsep}{2pt}
\renewcommand{\arraystretch}{1.0}
\resizebox{\linewidth}{!}{%
\begin{tabular}{@{}lcccc@{}}
\toprule
Policy & Coin & Nut Pick & Peg-in-Hole & Mean \\
\midrule
\textbf{\texttt{FEnc+FDec}} & \(43.6{\pm}8.3\) & \(83.7{\pm}6.1\) & \(18.2{\pm}7.5\) & \(48.5\) \\
\textbf{\texttt{NoEnc+FDec}} & \(48.6{\pm}8.3\) & \(80.7{\pm}11.3\) & \(26.6{\pm}7.6\) & \(52.0\) \\
\textbf{\texttt{NoEnc+GDec}} & \(54.9{\pm}7.5\) & \(\mathbf{88.9{\pm}4.1}\) & \(33.5{\pm}8.0\) & \(59.1\) \\
\textbf{\texttt{GEnc+FDec}} & \(\underline{60.2{\pm}9.1}\) & \(85.4{\pm}5.5\) & \(\underline{33.9{\pm}5.3}\) & \(\underline{59.8}\) \\
\textbf{\texttt{GEnc+GDec}} & \(\mathbf{65.6{\pm}4.8}\) & \(\underline{88.2{\pm}4.2}\) & \(\mathbf{43.8{\pm}4.2}\) & \(\mathbf{65.9}\) \\
\bottomrule
\end{tabular}
}
\vspace{-3mm}
\end{wraptable}

We fix the sensing input to \textbf{\texttt{W+FE}} and compare the policy interfaces defined in Sec.~\ref{subsec:policy_design}.

\textbf{O4: Group-structured modality fusion better preserves and exploits FingerEye’s feedback.}
\textbf{\texttt{GEnc+FDec}} improves over \textbf{\texttt{FEnc+FDec}} by \(11.3\) percentage points (Table~\ref{tab:sim_arch}), showing the benefit of forming group-level representations before fusion.
\textbf{\texttt{NoEnc+GDec}} improves over \textbf{\texttt{NoEnc+FDec}} by \(7.1\) percentage points, showing that grouped decoder access also matters.
Combining both gives the best result: \textbf{\texttt{GEnc+GDec}} reaches \(65.9\%\) mean SR, \(13.9\) percentage points above the strongest non-grouped baseline.
\ifdefined\arxivcombined
The decoder-attention diagnostic in Sec.~\ref{app:attention_diagnostics} shows substantial proprioceptive dominance in flat decoders; better-performing flat variants also preserve more \ours{} attention, consistent with the modality-competition interpretation.
\fi

\textbf{Real-world grouped-fusion comparison.}
On Coin Standing and Letter Retrieval, with all other settings fixed, \textbf{\texttt{GEnc+GDec}} achieves \(65.0\%\) mean SR versus \(52.5\%\) for the strongest non-grouped baseline, \textbf{\texttt{NoEnc+FDec}}. This \(12.5\)-percentage-point gain shows that group-structured fusion also benefits manipulation under real sensor noise. Additional wrench-input results are in Sec.~\ref{subsec:supp_wrench_input}.

\section{Limitations and Conclusions}
While \ours improves contact-rich dexterous manipulation, several limitations remain. The visually exposed acrylic surface remains low-friction despite high-friction material in camera-occluded regions; transparent high-friction coatings are a potential remedy. Normal-compression durability is evaluated in Sec.~\ref{subsec:supp_durability}, but longer-term and multi-axis durability remain untested. The camera placement and tag layout are manually designed; automatically optimizing sensor geometry could improve fingertip coverage, reduce blind regions, and better handle diverse contact configurations.
\ifdefined\arxivversion
Additional sensing-interface design trade-offs are discussed in Sec.~\ref{sec:design_tradeoff}.
\else
We discuss additional sensing-interface design trade-offs in the supplementary material.
\fi
Our experiments focus on single-arm, fixed-base manipulation, leaving mobile, bimanual, and humanoid settings for future work. Although our simulation supports controlled ablations and scalable evaluation, improving sim-to-real fidelity and incorporating reinforcement learning could enable larger-scale data generation, active sensing, and recovery beyond demonstrated behaviors.

Overall, \ours combines continuous binocular fingertip sensing and contact-induced deformation feedback with real-and-sim learning infrastructure and a group-structured policy interface. Across simulation and real-world tasks, this joint design improves contact-rich dexterous manipulation with diverse objects and interaction requirements.

\clearpage
\section*{Acknowledgement}

We thank Junyuan Cui for assistance with the simulation setup; Jinxuan Zhu, Jingxiang Guo, Zixuan Liu, Chongkai Gao, Chenrui Tie, Quantao Li, Yiwen Hou, and Junting Chen for their valuable discussions; and Yanhui Liu, Hengxu Yan, Zhijie Zhang, and Viking Wang for assistance with the hardware setup. We also thank Lihao Lin, Xueliang Xiong, and Juncheng Han for their assistance with the additional experiments conducted during the rebuttal. This research is supported by the Ministry of Education, Singapore, under the Academic Research Fund Tier 1 (FY2024). This work was also supported in part by the President's Graduate Fellowship and RoboScience. The views and conclusions expressed herein are those of the authors and should not be interpreted as necessarily representing the official policies or endorsements, either expressed or implied, of the sponsors.

{\small
\bibliography{references}
}

\clearpage
\ifdefined\arxivcombined
\else
\PassOptionsToPackage{numbers,sort&compress}{natbib}
\documentclass{article}
\usepackage{corl/corl_2026}
\usepackage{xcolor}

\definecolor{urlblue}{HTML}{006EE7}

\hypersetup{
    colorlinks=true,
    urlcolor=urlblue,
    linkcolor=urlblue,
    citecolor=urlblue,
    filecolor=urlblue,
}

\begin{document}
\raggedbottom
\title{\LARGE \bf
Supplementary Material for \ours: Learning Dexterous Manipulation with Continuous Vision-Tactile Sensing
}

\author{Anonymous Authors}

\maketitle
\fi

\renewcommand{\contentsname}{Appendix Contents}
\renewcommand{\thesubsection}{A.\arabic{subsection}}
\renewcommand{\thesubsubsection}{\thesubsection.\arabic{subsubsection}}
\newlength{\tocindentA} \setlength{\tocindentA}{0em}   
\newlength{\tocnumwidthA} \setlength{\tocnumwidthA}{1.5em} 
\newlength{\tocindentB} \setlength{\tocindentB}{1.5em} 
\newlength{\tocnumwidthB} \setlength{\tocnumwidthB}{2.2em} 
\newcommand{\MyTocLine}[5]{
  \par
  \begingroup
    \interlinepenalty10000 
    \parindent 0pt 
    \leftskip #1\relax 
    \rightskip 2.55em\relax 
    \parfillskip -\rightskip 
    \addtolength{\leftskip}{#2}
    \noindent \hspace{-#2}
    \makebox[#2][l]{#3}
    #4
    \nobreak\leaders\hbox{.}\hfill #5
    \par
    \vspace{0.6ex}
  \endgroup
}
\etocsetstyle{subsection}
  {\addvspace{0.5ex}} 
  {} 
  {%
    \bfseries 
    \MyTocLine{\tocindentA}{\tocnumwidthA}{\etocnumber}{\etocname}{\etocpage}%
  }
  {}
\etocsetstyle{subsubsection}
  {} 
  {} 
  {
    \mdseries 
    \MyTocLine{\tocindentB}{\tocnumwidthB}{\etocnumber}{\etocname}{\etocpage}%
  }
  {}
\appendix
\localtableofcontents 

\subsection{Additional Sensing Interface Design Details}

\subsubsection{Hardware Design Details}
\label{subsec:supp_hardware_design}
This section expands the implementation details behind the sensing-interface design highlights in the main paper.
Each \ours module integrates a binocular RGB camera system, a transparent acrylic plate with an AprilTag array, and a compliant silicone ring that converts contact into measurable plate-pose changes.
The main paper shows the overall geometry and exploded view; here we specify the mechanical and optical choices that are shortened there.

\begin{wrapfigure}[9]{r}{0.46\linewidth}
\vspace{-5mm}
\centering
\includegraphics[width=\linewidth]{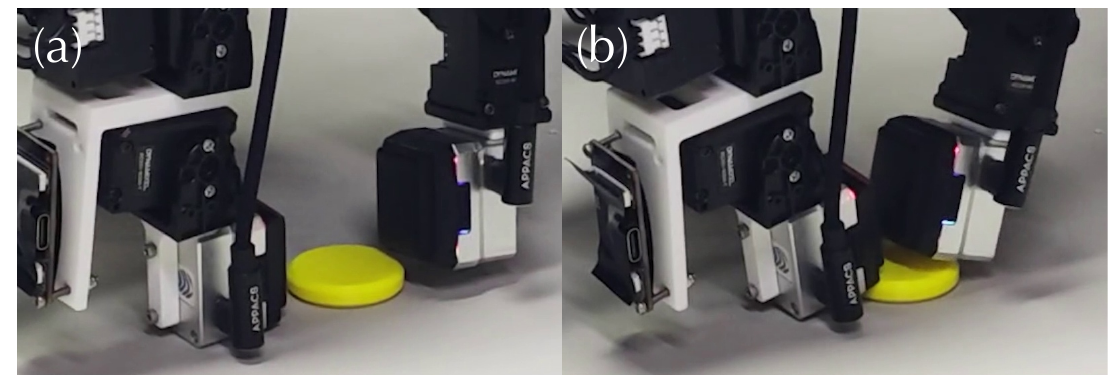}
\caption{\textbf{GelSight failure in coin standing.}
The square sensor geometry makes edge wedging difficult in this setup.}
\label{fig:gelsight}
\vspace{-5mm}
\end{wrapfigure}

\vspace{1mm}
\noindent\textbf{Wedge-shaped fingertip module.}
The assembled module has a compact wedge-shaped form factor with an overall size of approximately \(41.7\times30.0\times25.0\)~\si{mm}.
This geometry is chosen so the sensor can be mounted at dexterous fingertips while still entering narrow spaces, approaching object edges, and wedging under thin objects such as coins or letters.
The wedge shape is therefore not only a packaging choice: it determines whether the fingertip can physically reach the contact configurations used in edge engagement and contact initiation tasks, as also illustrated by the GelSight wedging failure in Fig.~\ref{fig:gelsight}.

\vspace{1mm}
\noindent\textbf{Complementary binocular cameras.}
Each module uses two HBVCAM-2307-FPC82 V11 USB camera modules, each approximately \(8.6\times8.6\times5.4\)~\si{mm} with a \(120^{\circ}\) field of view.
The tip camera is mounted approximately perpendicular to the acrylic plate at a short working distance of \(\SI{10}{mm}\), giving a sharp view of the AprilTag array and nearby contact surface.
The root camera is mounted closer to the finger root and tilted by \(21.48^{\circ}\) relative to the fingertip horizontal direction, with a longer working distance of roughly \(\SI{80}{mm}\).
Together, the two views cover different interaction phases: the tip camera emphasizes local deformation and close-range geometry, while the root camera preserves a wider scene view during approach and manipulation.
Their spatially offset viewpoints also provide implicit depth cues that help disambiguate close-range object position and contact alignment.

\vspace{1mm}
\noindent\textbf{Compliant peripheral ring.}
The soft ring is fabricated by silicone molding and mechanically bonded around the transparent acrylic cover.
Its internal triangular elements provide directional compliance: the ring can deform under external forces and torques while preserving enough shape stability for repeatable pose estimation.
Because the compliant region surrounds the plate rather than occupying the central visual path, frontal contact, lateral contact, and peripheral loading can all induce measurable acrylic-plate motion without blocking the RGB view.
This is the mechanical basis for using plate-pose change as a contact-wrench proxy.

\vspace{1mm}
\noindent\textbf{Acrylic cover and AprilTag array.}
A \(\SI{1}{mm}\)-thick transparent acrylic plate protects the optical path and carries the fiducial pattern on its inner surface.
The current tag layout uses \(35\) tags from the \texttt{tag25h9} family arranged as a compact central cluster surrounded by a ring of tags for broad spatial coverage.
Each tag is \(\SI{2}{mm}\) wide and includes a \(\SI{0.2}{mm}\) white border for reliable detection.
This distributed layout is designed for partial visibility: when force, occlusion, or local deformation hides some tags, the remaining visible corners still constrain one global plate-pose estimate.

\vspace{1mm}
\noindent\textbf{Mechanical assembly.}
The camera wrappers, sensor base, soft-ring molds, and supporting fixtures are 3D printed, while M2\(\times\)4 screws fix the camera wrappers and help preserve optical alignment.
The fully assembled module connects to a host PC through USB-C camera interfaces.
The fabrication procedure and bill of materials are given in Sec.~\ref{subsec:supp_fabrication}.

\subsubsection{Fabrication Details and Bill of Materials}
\label{subsec:supp_fabrication}
These details support the main-paper claim that \ours can be built from low-cost off-the-shelf and 3D-printed components.

The fabrication process consists of five steps.
\begin{enumerate}[leftmargin=5mm, topsep=1pt, itemsep=1pt]
    \item \textbf{3D printing.}
    Print the sensor base, camera wrappers, and mold components, including the lower mold, upper mold, and core insert.
    \item \textbf{Silicone preparation and casting.}
    Mix the silicone base with curing agent at a mass ratio of \(\text{base}:\text{curing agent}=100:(1\sim3)\).
    Pour the mixture into the lower mold, assemble the upper mold and core insert, press the assembly onto the lower mold, and cure for more than \(4\) hours.
    After curing, remove the mold components and demold the soft silicone ring.
    \item \textbf{Acrylic plate and tag preparation.}
    Apply the AprilTag sticker to the inner surface of the acrylic plate, bond the soft silicone ring to the sticker using silicone adhesive (JL-401), and let the adhesive fully cure.
    \item \textbf{Camera assembly.}
    Secure the camera wrappers to the sensor base using M2\(\times\)4 screws and install the camera modules into the wrappers.
    \item \textbf{Wiring and connection.}
    Connect the assembled module to a host PC using a USB-C cable.
\end{enumerate}

\begin{figure}[H]
\vspace{-1mm}
\centering
\includegraphics[width=0.92\linewidth]{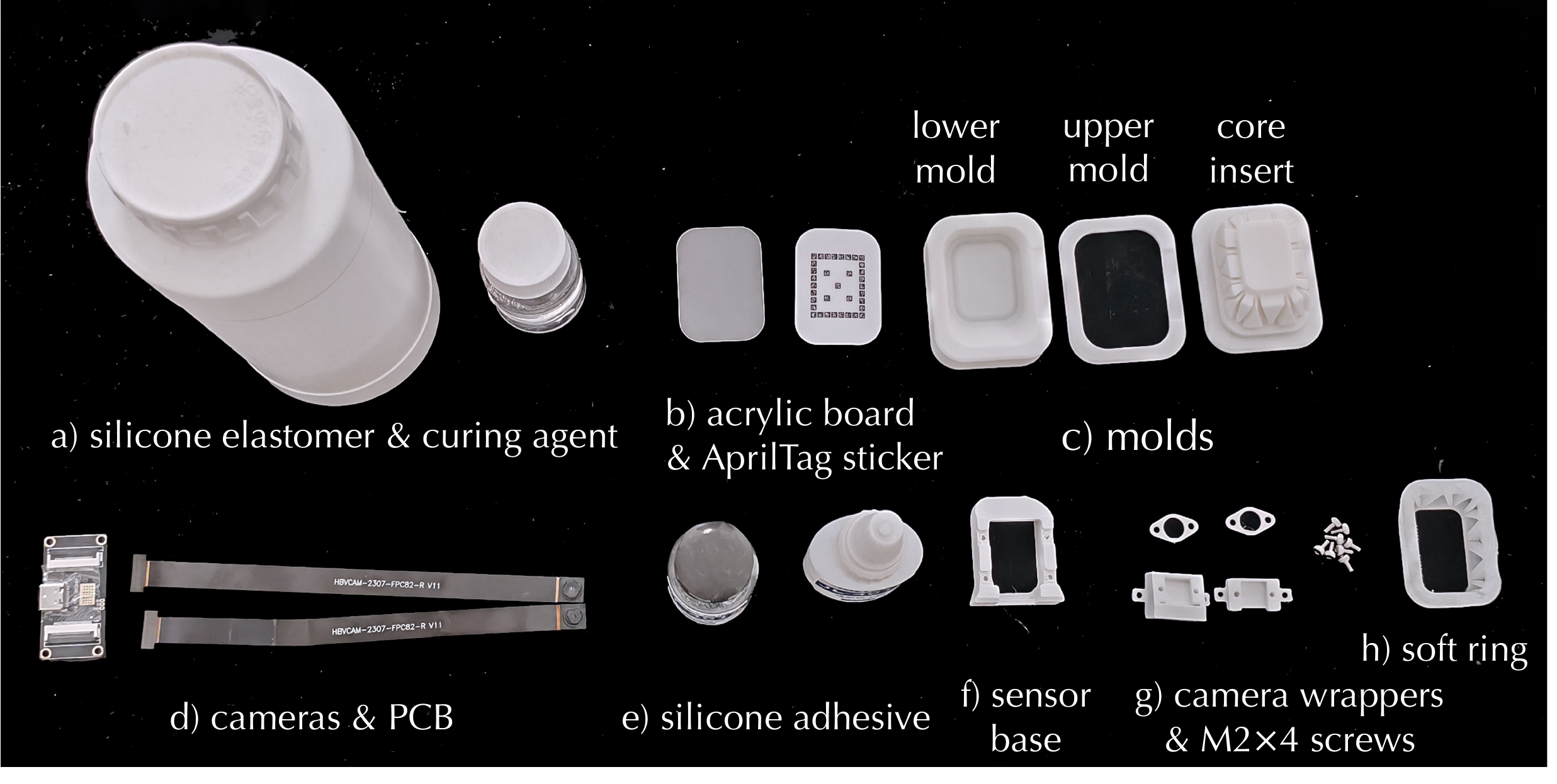}
\caption{\textbf{Materials used for \ours fabrication.}
The module uses silicone elastomer and curing agent, camera modules, an acrylic plate with AprilTag sticker, silicone adhesive, and 3D-printed mold and fixture components.}
\label{fig:materials_for_Fingers}
\vspace{-3mm}
\end{figure}

\begin{table}[H]
\centering
\caption{Bill of materials for one \ours module.}
\label{tab:bill}
\small
\setlength{\tabcolsep}{7pt}
\begin{tabular}{lc}
\toprule
Item & Cost (USD) \\
\midrule
Acrylic plate & 0.28 \\
AprilTag sticker & 0.42 \\
Silicone elastomer \& curing agent (20\,g) & 0.10 \\
Silicone adhesive (JL-401, \(\sim\)0.5\,ml) & 0.17 \\
Soft silicone ring & 0.02 \\
Sensor base (3D-printed) & 0.44 \\
Camera wrappers (3D-printed) & 0.45 \\
Soft-ring mold (3D-printed) & 1.78 \\
M2\(\times\)4 screws (8 pcs) & 0.04 \\
Camera modules (\(\times\)2) & 56.49 \\
\midrule
Total & 60.19 \\
\bottomrule
\end{tabular}
\vspace{-3mm}
\end{table}

\subsubsection{Multi-Tag Deformation Estimation}
\label{subsec:appendix_marker}
This section gives the deformation estimator used by the multi-tag pose tracking design.
The AprilTag array is rigidly attached to the acrylic plate, so the estimated plate pose provides a compact proxy for deformation of the surrounding compliant ring.
For each detected tag, we extract four image corners and aggregate all visible corners into one multi-tag correspondence set,
\[
\big\{(\mathbf{X}_{i,j}^{\mathrm{ref}},\mathbf{u}_{i,j})\big\},
\]
where \(\mathbf{X}_{i,j}^{\mathrm{ref}}\) is the 3D corner in the reference-tag frame and \(\mathbf{u}_{i,j}\) is the observed image corner.
We estimate the acrylic-plate pose by solving
\begin{equation}
\mathbf{T}_{\mathrm{cam}\leftarrow\mathrm{ref}}
=
\arg\min_{\mathbf{T}\in SE(3)}
\sum_{i,j}
\left\|
\mathbf{u}_{i,j}
-
\pi\!\left(\mathbf{T}\mathbf{X}_{i,j}^{\mathrm{ref}}\right)
\right\|_2^2,
\end{equation}
using EPnP followed by Levenberg--Marquardt refinement.
An alternative is to solve a pose for each visible tag independently and average the resulting tag poses.
We found this averaging strategy less stable because each tag estimate has its own local corner noise, occlusion pattern, and degeneracy, so a single poorly localized tag can perturb the averaged pose.
The joint multi-tag solve is more stable because all visible corners are optimized under one shared rigid-plate transform, making the estimate overdetermined and reducing jitter when individual tags are occluded, blurred, or locally distorted.
During interaction, we subtract the no-contact reference pose from the current plate pose to obtain the six-dimensional deformation vector used in the wrench calibration in Sec.~\ref{subsec:supp_corr_setup}.

\subsubsection{Comparison with GelSight and Keyline Tracking}
\label{subsec:supp_gelsight}
This section clarifies how the design differs from central-contact gel sensors and keyline-marker STS tracking.
\par\smallskip
We attempted to use GelSight for the \textit{coin standing} task, but encountered two practical issues during teleoperation data collection.
First, the sensor did not provide a useful signal for table-edge interaction because the relevant contact occurred near the side of the fingertip rather than on the central gel surface.
Second, the square sensor geometry made it difficult to wedge under the coin smoothly.
In contrast, \ours uses a wedge-shaped fingertip with a peripheral soft ring, allowing the robot to engage the coin edge while still obtaining visual and deformation cues from the fingertip.

\par\smallskip
\begin{wrapfigure}[12]{r}{0.45\linewidth}
\vspace{-4mm}
\centering
\includegraphics[width=\linewidth]{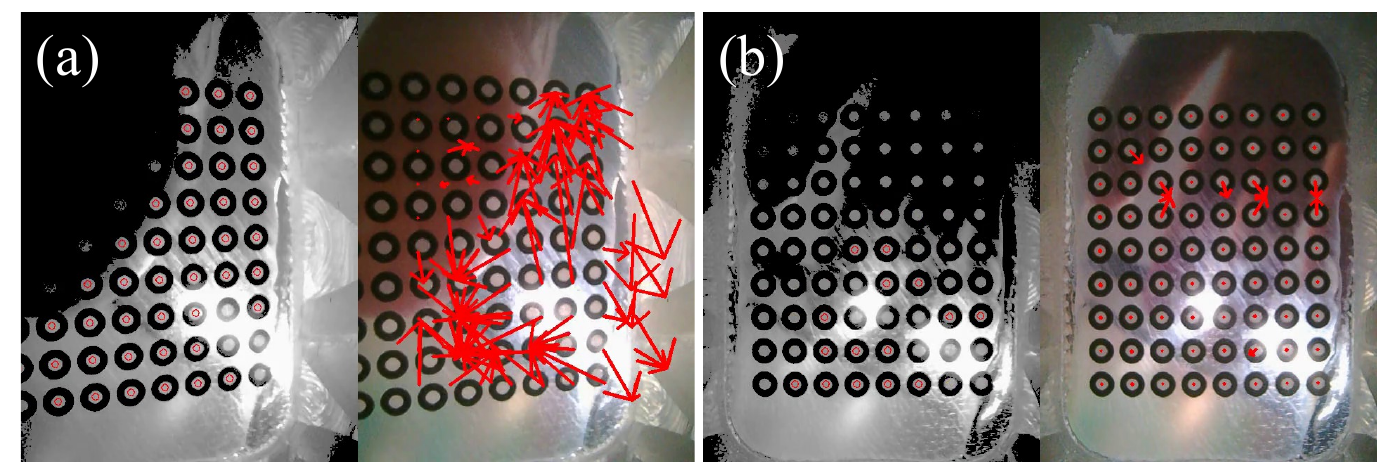}
\vspace{-2mm}
\caption{\textbf{Qualitative comparison of marker-tracking pipelines.}
Keyline tracking is brittle under deformation and lighting, while AprilTags preserve tag identity and multi-corner pose constraints.}
\label{fig:keyline_comparison}
\vspace{-3mm}
\end{wrapfigure}
We also compared the AprilTag-based pipeline with a keyline-marker pipeline inspired by TacThru~\cite{li2026simultaneous}.
The keyline pipeline detects circular markers with blob detection, then relies on nearest-neighbor association and temporal filtering.
In our setup, it was less stable when deformation between frames was large or when ambient illumination varied, because both blob responses and marker association became brittle.
By contrast, explicit tag identities and multi-corner constraints provide better conditioning for the unified PnP solve.

\subsubsection{Design Trade-offs}
\label{sec:design_tradeoff}

\vspace{1mm}
\noindent\textbf{Complementary binocular views vs. a single compact camera.}
A single fingertip camera would reduce wiring and calibration complexity, but it would force one optical configuration to cover both the close-range tag array and the wider pre-contact scene.
\ours instead separates these roles across root and tip cameras.
This increases hardware complexity slightly, but it provides the local coverage and close-range depth cues evaluated by the monocular-vs.-binocular ablations in the main paper.

\vspace{1mm}
\noindent\textbf{Transparency--friction trade-off.}
The acrylic cover improves optical clarity and keeps the RGB stream stable across pre-contact and post-contact phases, but a rigid surface can reduce conformity and friction compared with soft gel interfaces.
We mitigate this by placing high-friction gripping material on regions not visible to the cameras, such as perimeter regions and areas hidden behind tags, to improve grasp reliability without blocking the visual sensing path. The visually exposed surface remains low-friction; transparent high-friction coatings are a potential future improvement.

\vspace{1mm}
\noindent\textbf{Passive lighting vs. active self-illumination.}
We avoid active or colored self-illumination because the same RGB observations are used for both sensing and policy learning.
Passive lighting keeps image appearance close to ordinary visual observations, while contrast enhancement improves AprilTag detection under low-light conditions without introducing illumination-specific color artifacts.
The lighting-variation robustness result in the main paper further supports this choice: stable tag-array pose estimation does not require switching to active colored illumination.

\vspace{1mm}
\noindent\textbf{Peripheral compliance for unobstructed contact sensing.}
Unlike designs that place a deformable gel in the central optical path, \ours keeps the central view clear and places compliance around the plate perimeter.
This peripheral compliant interface allows frontal contact, edge engagement, and side loading to induce measurable plate-pose changes.
The expanded deformation region is important for behaviors such as wedging, sliding, and reorientation, where useful contact often occurs near the fingertip edge.

\vspace{1mm}
\noindent\textbf{Continuous RGB sensing instead of explicit contact-map reconstruction.}
\ours does not explicitly reconstruct dense contact shape or contact location from the real sensor stream.
This trades explicit contact-map output for visual clarity, easy fabrication, continuous pre-contact perception, implicit depth from binocular RGB, and marker-tracked post-contact deformation in one sensing stream.
The resulting output stays simple for policy learning while preserving the contact-sensitive cues needed by the evaluated tasks.

\subsubsection{Tag-Count Ablation}
\label{subsec:supp_tag_count}
We evaluate nested \(1/5/20/35\)-tag layouts with the same PnP pipeline. Table~\ref{tab:tag_count} reports static pose jitter under lighting and background variations, the mean number of detected tags during motion, and deviation from the simultaneous \(35\)-tag estimate. This deviation measures agreement with the full-layout estimate, rather than absolute pose error against external ground truth.
\begin{table}[H]
\centering
\caption{Pose robustness versus tag count. Lower jitter and deviation are better.}
\label{tab:tag_count}
\small
\setlength{\tabcolsep}{8pt}
\begin{tabular}{lccc}
\toprule
Tags & Jitter (mm/$^{\circ}$) & Mean detected tags & Deviation (mm/$^{\circ}$) \\
\midrule
1 & 0.426 / 6.969 & 0.85 & 0.363 / 6.823 \\
5 & 0.207 / 1.473 & 3.65 & 0.237 / 2.203 \\
20 & 0.015 / 0.120 & 11.86 & 0.068 / 0.326 \\
35 & 0.012 / 0.109 & 19.82 & Reference \\
\bottomrule
\end{tabular}
\end{table}
In a simulation ablation with the same trajectories and policy setup, the \(20\)-tag layout achieves \(68.3\%\) mean SR versus \(65.9\%\) for \(35\) tags. Together, these results suggest a trade-off between visual coverage and pose stability: more tags improve tracking redundancy, but do not necessarily improve policy success.

\subsection{Sensing Interface Capability and Experiment Details}

\subsubsection{Wrench--Deformation Calibration Setup}
\label{subsec:supp_corr_setup}
This section gives the calibration protocol behind the main-paper claim that plate-pose change is a contact-wrench proxy.
\par\smallskip
\begin{wrapfigure}[17]{r}{0.40\linewidth}
\vspace{-2mm}
\centering
\includegraphics[width=\linewidth]{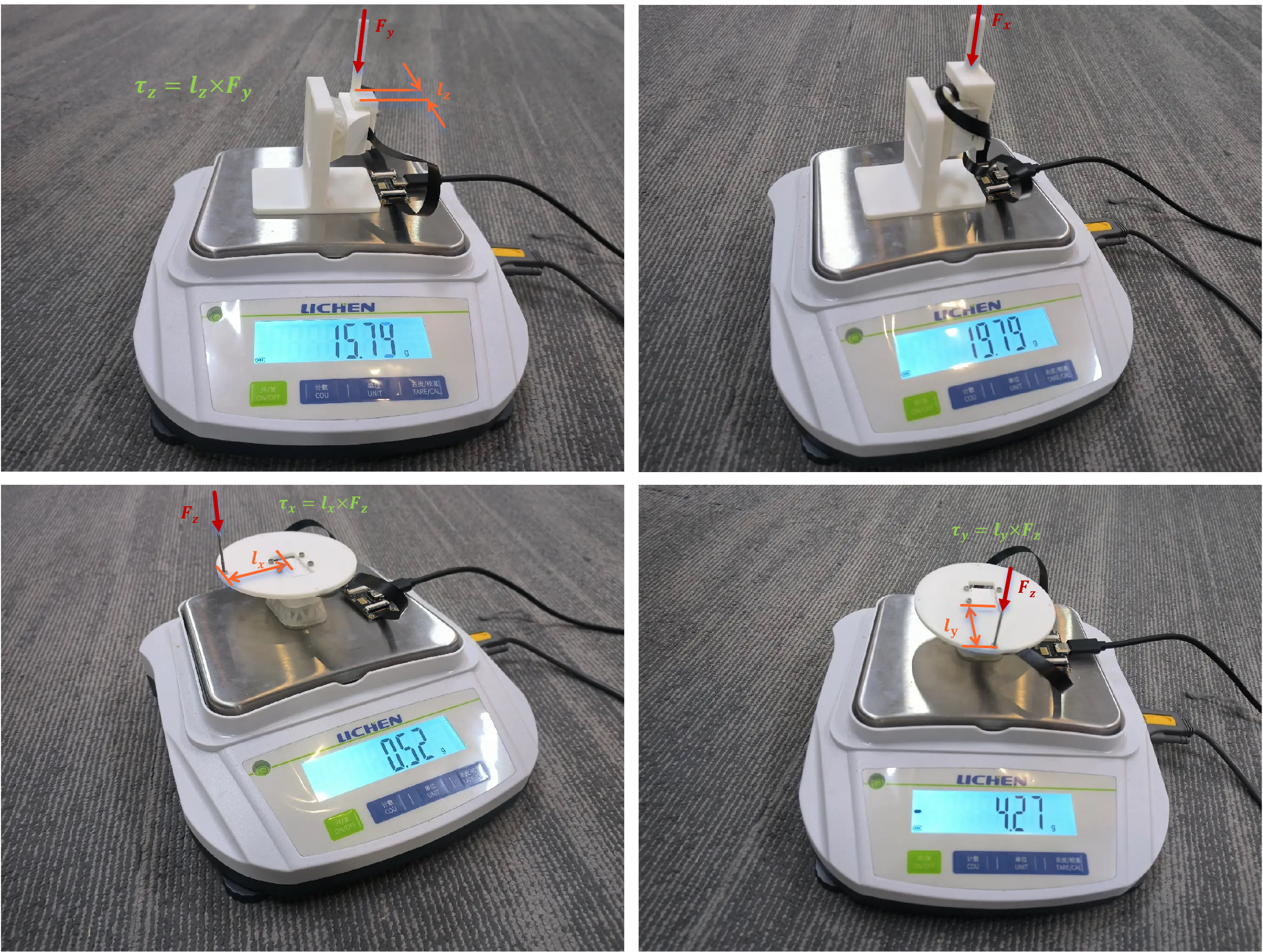}
\vspace{-1mm}
\caption{\textbf{Calibration setup.}
A scale and 3D-printed fixtures provide axis-specific force and torque loading while the tag-array pose is recorded.}
\label{fig:setup}
\vspace{0mm}
\end{wrapfigure}
Let \(\boldsymbol{\Delta L}\) denote the six-dimensional pose change of the acrylic plate relative to a no-contact reference,
\[
\boldsymbol{\Delta L}
=
[\delta l_x,\delta l_y,\delta l_z,\delta \theta_x,\delta \theta_y,\delta \theta_z]^{\mathsf{T}},
\]
with translations in millimeters and rotations in radians.
Let \(\mathbf{F}\) denote the applied force and torque,
\[
\mathbf{F}
=
[f_x,f_y,f_z,\tau_x,\tau_y,\tau_z]^{\mathsf{T}} .
\]
We fit a mapping \(g(\cdot)\) such that \(\mathbf{F}=g(\boldsymbol{\Delta L})\).
For force measurements, a digital scale measures the applied load while 3D-printed fixtures constrain the loading direction and contact location.
For torque measurements, a force is applied at a known lever arm, and torque is computed from the measured force and lever-arm length.
During each trial, forces or torques are applied along one axis at a time while the acrylic-plate pose is estimated from AprilTag tracking.
We collect over \(1{,}000\) synchronized wrench--pose pairs spanning all six translational and rotational degrees of freedom.
For each axis, \(80\%\) of the samples are used for fitting and \(20\%\) are held out for testing.

\par\vspace{-1mm}
\begin{wrapfigure}[10]{r}{0.45\linewidth}
\vspace{-6mm}
\centering
\includegraphics[width=\linewidth]{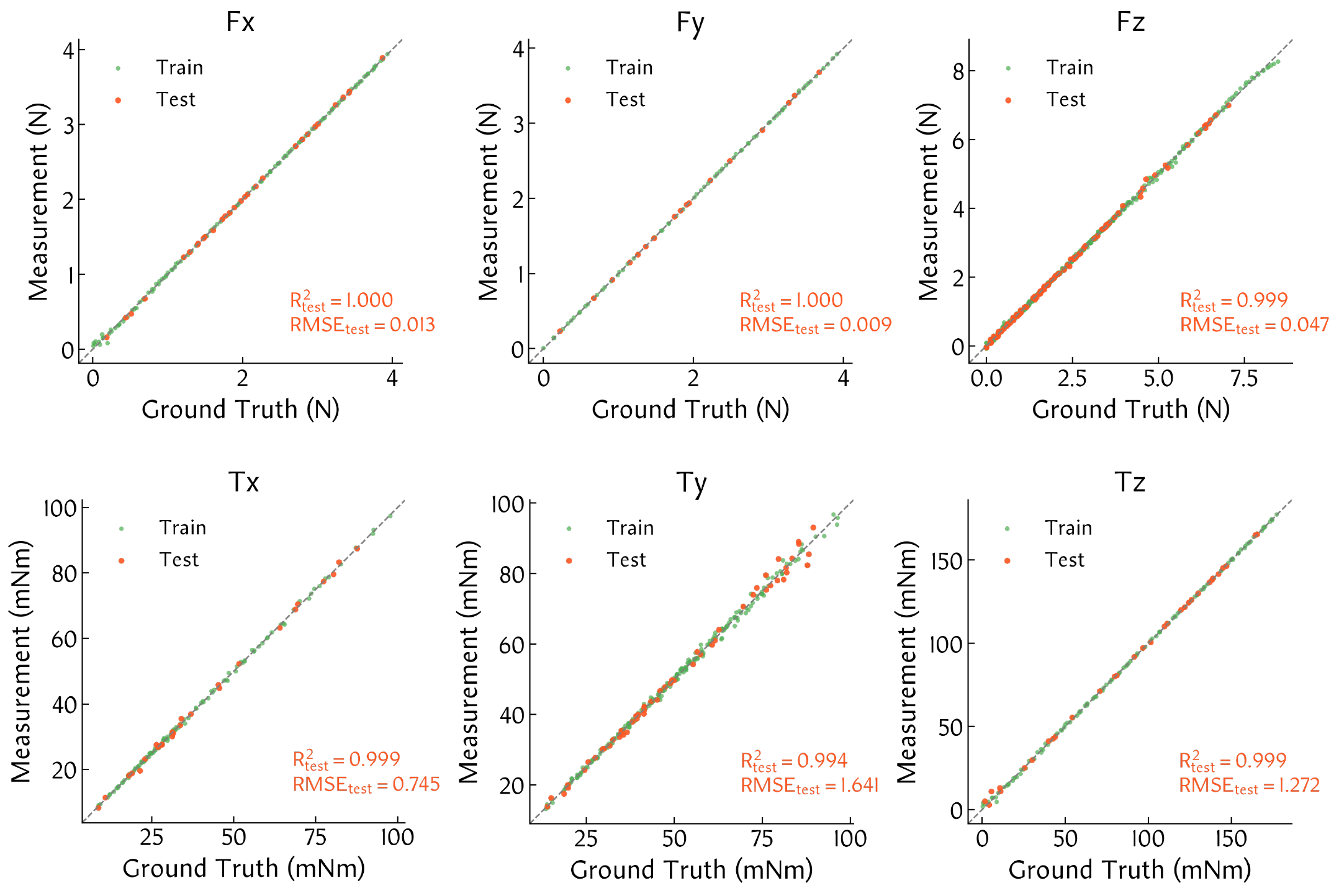}
\vspace{-2mm}
\caption{\textbf{Wrench--deformation calibration.}
Predicted forces and torques closely match held-out measurements.}
\label{fig:hw_exp}
\vspace{-3mm}
\end{wrapfigure}
Fig.~\ref{fig:hw_exp} shows that the learned wrench--deformation mapping is close to linear and generalizes consistently across held-out samples.
The three force components achieve \(R^2_{\rm test}\geq0.999\), with held-out RMSE values of \(0.013\), \(0.009\), and \(0.047\)~N for \(F_x\), \(F_y\), and \(F_z\), respectively.
The torque components also remain well predicted, with \(R^2_{\rm test}=0.999\), \(0.994\), and \(0.999\) for \(\tau_x\), \(\tau_y\), and \(\tau_z\).
These results support the use of acrylic-plate pose change as a compact contact-wrench proxy for the sensitivity analysis in Sec.~\ref{subsec:supp_sensitivity}.

\subsubsection{Sensitivity Analysis Details}
\label{subsec:supp_sensitivity}
This section derives the minimum detectable wrench reported in the main paper.
Following fiducial-based sensitivity analysis in prior work~\cite{ouyang2020low,zhu2025shapeforce}, we estimate the minimum detectable pose change of \ours from pixel-level localization accuracy and propagate it through the calibrated wrench--deformation model.

\vspace{1mm}
\noindent\textbf{Subpixel localization assumption.}
We assume a subpixel localization accuracy of \(d_R=0.25\)~px for AprilTag detection.

\vspace{1mm}
\noindent\textbf{Translational sensitivity.}
Let \(w_{\text{tag}}\) denote the physical width of the AprilTag and \(w_{\text{img}}\) denote its observed image width.
A displacement of \(d_R\) pixels corresponds to a minimum detectable translation
\begin{equation}
\Delta l_{\min}
=
\frac{w_{\text{tag}}}{w_{\text{img}}}\,d_R .
\end{equation}
Using \(w_{\text{tag}}=2.0\)~mm and \(w_{\text{img}}=37\)~px gives
\[
\Delta l_{\min}\approx0.0135~\mathrm{mm}.
\]

\vspace{1mm}
\noindent\textbf{Rotational sensitivity.}
For a reference rotation angle \(\theta\), the corresponding pixel displacement at radius \(r\), the half-width of the observed tag image patch, is given by the chord length \(2r\sin(\theta/2)\).
A \(d_R\)-pixel displacement therefore corresponds to a minimum detectable rotation
\begin{equation}
\Delta \theta_{\min}
=
\frac{\theta}{2r\sin(\theta/2)}\,d_R .
\end{equation}
Using \(\theta=\pi/12\) and \(r=18.5\)~px gives
\[
\Delta \theta_{\min}\approx0.0136~\mathrm{rad}.
\]

\vspace{1mm}
\noindent\textbf{Pose sensitivity summary.}
The resulting minimum detectable 6-DoF pose change is
\begin{equation}
\boldsymbol{\Delta L}_{\min}
=
[\,0.0135,\;0.0135,\;0.0135,\;0.0136,\;0.0136,\;0.0136\,]^{\mathsf{T}},
\end{equation}
where the first three entries are in millimeters and the last three are in radians.

\vspace{1mm}
\noindent\textbf{Propagation to wrench sensitivity.}
Propagating \(\boldsymbol{\Delta L}_{\min}\) through the calibrated linear wrench--deformation model yields
\begin{equation}
\mathbf{F}_{\min}
=
[\,4.30,\;4.22,\;9.93,\;0.32,\;0.13,\;8.55\,]^{\mathsf{T}},
\end{equation}
where the first three entries correspond to force sensitivity in mN and the last three entries correspond to torque sensitivity in mN\(\!\cdot\!\)m.

\subsubsection{Delicate Grasping Setup and Hyperparameters}
\label{subsec:supp_delicate_setup}
This section gives the control details behind the contact-onset utility experiment in the main paper.
Inspired by contact-sensitive grasping studies such as SpikeAtac~\cite{chang2025spikeatac}, we evaluate whether \ours can trigger gentle stopping and lifting behavior on fragile or deformable objects.
All grasping motions follow predefined joint-space trajectories generated by linear interpolation between an initial pose and a target pose.
Joint commands are issued at a fixed control interval of \(0.02\)~s per frame.
To detect contact, we monitor the acrylic-plate normal displacement \(\Delta z\) estimated from AprilTag tracking.
When \(\Delta z\) exceeds the object-specific threshold in Table~\ref{tab:delicate_grasping_hyperparams}, the approach motion stops and the robot lifts the object.
Most objects are placed on a custom 3D-printed support platform so that thin or deformable objects can be contacted without scraping the table.

\begin{table}[H]
\centering
\caption{Hyperparameters for delicate grasping experiments.}
\label{tab:delicate_grasping_hyperparams}
    \small
    \setlength{\tabcolsep}{7pt}
    \begin{tabular}{lcc}
        \toprule
        Object & Contact threshold (mm) & Total frames \\
        \midrule
        0.7\,mm pencil lead & 0.02 & 15 \\
        Chip & 0.10 & 30 \\
        Dessert cup & 0.05 & 10 \\
        Seaweed & 0.01 & 150 \\
        Grape & 0.06 & 10 \\
        Cone & 0.10 & 20 \\
        Eggshell & 0.05 & 20 \\
        Wafer & 0.05 & 5 \\
        Balloon & 0.05 & 3 \\
        \bottomrule
    \end{tabular}
    \vspace{-3mm}
\end{table}

\subsubsection{Normal-Compression Durability}
\label{subsec:supp_durability}
After \(10{,}000\) normal-compression cycles at \(3\) mm and \(1\) Hz, we observe no cracking or delamination. Before and after cycling, we record scale-measured normal force and estimated plate displacement \(\Delta z\) over quasi-static loading--unloading sweeps. At matched force levels, the maximum displacement hysteresis (the largest loading--unloading difference in \(\Delta z\)) increases from \(0.39\) to \(0.46\) mm, or approximately \(13\%\) to \(15\%\) of the \(3\)-mm deformation range. The pre- and post-cycling displacement responses differ by \(0.13\) mm RMS. Reevaluation with the cycled sensor on Coin Standing and Chip Picking (\(20\) trials each) leaves Coin SR unchanged and yields one fewer successful Chip trial. These tests suggest largely retained performance under normal compression; longer-term and multi-axis durability remain to be evaluated.

\subsection{Learning Interface Details}

\subsubsection{Policy Architecture Details}
\label{sec:supp_policy_arch}
\begin{figure}[H]
\centering
\includegraphics[width=\textwidth]{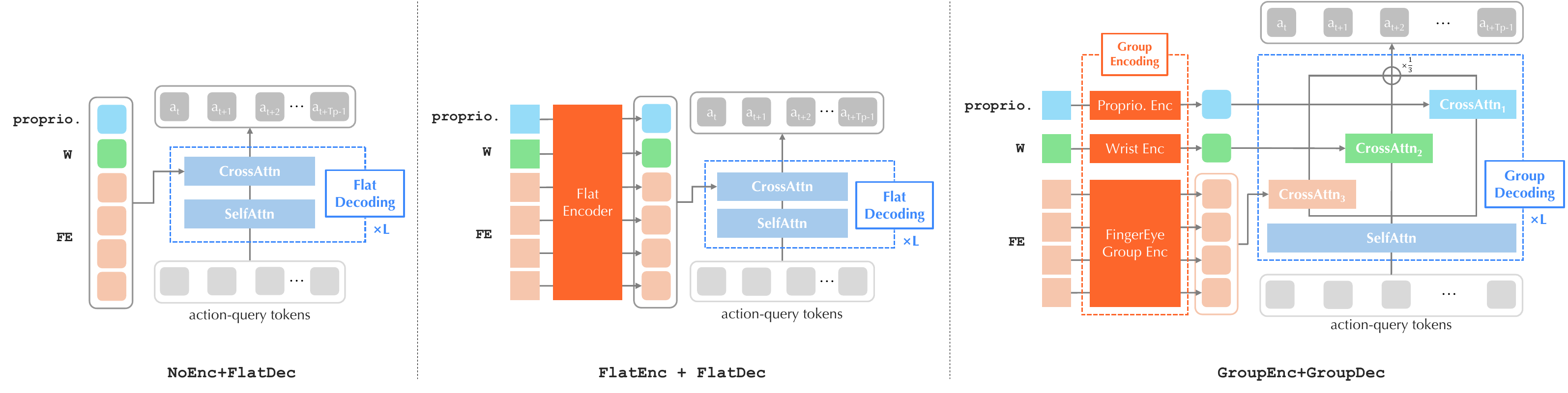}
\caption{\textbf{Policy architecture comparison.}
The ablation varies how observation tokens are encoded and how action-query tokens attend to modality groups; \textbf{\texttt{FEnc}}/\textbf{\texttt{GEnc}} and \textbf{\texttt{FDec}}/\textbf{\texttt{GDec}} abbreviate the \textbf{\texttt{FlatEnc}}/\textbf{\texttt{GroupEnc}} and \textbf{\texttt{FlatDec}}/\textbf{\texttt{GroupDec}} labels shown in the diagram.
Left: \textbf{\texttt{NoEnc+FDec}} exposes tokenized observations directly to a flat decoder.
Middle: \textbf{\texttt{FEnc+FDec}} applies flat self-attention across all observation tokens before flat decoding.
Right: \textbf{\texttt{GEnc+GDec}} encodes proprioception, wrist, and \ours groups separately, then gives each group an independent decoder cross-attention path.}
\label{fig:nn_arch}
\end{figure}

This section gives the tensor-level policy architecture used for the current policy-interface comparisons.
It specifies the temporal structure, tokenization, self-attention encoder variants, decoder cross-attention variants, and action projection used in the reported ablations.
Fig.~\ref{fig:nn_arch} visualizes representative encoder--decoder variants before the tensor-level specification below.

\vspace{1mm}
\noindent\textbf{Receding-horizon action chunking.}
At each control step \(t\), the policy conditions on the most recent \(T_o=1\) observation and predicts an action chunk of length \(T_p\).
The robot executes the first \(T_a\) actions and then replans in a receding-horizon manner, following action-chunking policies~\cite{zhao2023act,chi2025diffusion}.
The predicted action chunk is
\begin{equation}
\bm{A}_t
=
\pi_{\theta}(\bm{O}_t)
=
(\bm{a}_t,\bm{a}_{t+1},\ldots,\bm{a}_{t+T_p-1}),
\end{equation}
where each \(\bm{a}_t\in\mathbb{R}^{d_a}\) lies in the task-specific action space listed in Table~\ref{tab:task_policy_settings}.

\vspace{1mm}
\noindent\textbf{Observation groups.}
We write the observation as a tuple of modality groups
\begin{equation}
\bm{O}_t=(\bm{o}_t^1,\ldots,\bm{o}_t^M),
\end{equation}
where groups include wrist RGB, robot proprioception, \ours RGB, and optional auxiliary groups such as plate-pose histories or post-contact tactile-map tokens.
All non-action inputs are represented as tokens before encoding; optional plate-pose histories are therefore treated as additional observation tokens, not as query modulation terms.
RGB images are resized to \(192\times256\), and the proprioceptive state is projected from \(\bm{J}_t\in\mathbb{R}^{d_s}\).

\vspace{1mm}
\noindent\textbf{Visual tokenization with cached RADIO summaries.}
Let \(i\in\{1,\ldots,N_{\mathrm{cam}}\}\) index the wrist and fingertip cameras.
Each RGB image \(\bm{I}_t^{(i)}\) is passed independently through a frozen RADIO image backbone~\cite{ranzinger2024radio}.
The backbone outputs one global summary vector
\begin{equation}
\bm{s}_t^{(i)}\in\mathbb{R}^{2304}.
\end{equation}
For efficient training, these summary vectors are precomputed and cached offline.
During policy training, each camera summary is projected to the policy dimension \(d=512\) by a camera-specific MLP
\begin{equation}
\phi_i:\mathbb{R}^{2304}\rightarrow\mathbb{R}^{512},\qquad
\bm{v}_t^{(i)}=\phi_i(\bm{s}_t^{(i)}).
\end{equation}
Projection MLPs are implemented as two-layer MLPs with GELU activations.
We add a learnable camera-identity embedding \(\bm{e}_{\mathrm{cam}}^{(i)}\in\mathbb{R}^{512}\) to form the final visual token
\begin{equation}
\bm{u}_t^{(i)}
=
\bm{v}_t^{(i)}+\bm{e}_{\mathrm{cam}}^{(i)}.
\end{equation}
Vector observations, such as proprioception and optional plate-pose histories, are flattened and projected by group-specific MLPs into the same \(512\)-dimensional token space.
The \ours RGB group contains the projected tokens from the fingertip-mounted cameras, with camera-identity embeddings preserving finger and viewpoint identity.
The resulting token sets are grouped as
\begin{equation}
\bm{U}_t^m=\tau_m(\bm{o}_t^m),\qquad
\bm{U}_t=[\bm{U}_t^1,\ldots,\bm{U}_t^M],
\end{equation}
where \(\tau_m\) denotes the tokenizer for modality group \(m\).

\vspace{1mm}
\noindent\textbf{Encoder-interface variants.}
The encoder ablation changes the self-attention range before decoding.
\textbf{\texttt{NoEnc}} directly exposes tokenized observations to the decoder without pre-decoder self-attention.
\textbf{\texttt{FlatEnc}} applies layers of Transformer self-attention~\cite{vaswani2017attention} over all tokens together, so wrist, proprioception, \ours, and optional auxiliary tokens can attend to each other immediately.
\textbf{\texttt{GroupEnc}} applies self-attention within each modality group only when the group contains at least two tokens; in particular, all \ours tokens can attend to other \ours tokens, but they do not attend to wrist or proprioceptive tokens until decoder fusion.
Formally,
\begin{equation}
\begin{array}{@{}r@{\;}c@{\quad}r@{\;}c@{\;}l@{}}
\text{\normalfont\bfseries\ttfamily NoEnc}
&:&
\bm{Z}_t^{\mathrm{noenc}}
&=&
\bm{U}_t, \\
\text{\normalfont\bfseries\ttfamily FlatEnc}
&:&
\bm{Z}_t^{\mathrm{flat}}
&=&
E_{\mathrm{flat}}(\bm{U}_t), \\
\text{\normalfont\bfseries\ttfamily GroupEnc}
&:&
\bm{Z}_t^{\mathrm{group}}
&=&
[E_1(\bm{U}_t^1),\ldots,E_M(\bm{U}_t^M)] .
\end{array}
\end{equation}
where \(E_{\mathrm{flat}}\) is a stack of self-attention and feed-forward layers over all tokens, and \(E_m\) is a group-local stack when \(|\bm{U}_t^m|\geq 2\) and the identity mapping when \(|\bm{U}_t^m|<2\).
The default Transformer stacks use \(4\) layers, \(16\) attention heads, hidden dimension \(512\), and feed-forward dimension \(2048\); size-matched architecture ablations adjust hidden or feed-forward widths only when needed.
For \ours, \textbf{\texttt{GroupEnc}} lets fingertip-local views integrate contact geometry before competing with global wrist or proprioceptive cues.

\vspace{1mm}
\noindent\textbf{Action-query initialization.}
The decoder starts from clean learnable action-query embeddings:
\begin{equation}
\bm{h}_{t,k}^{0}=\bm{q}_k,\qquad
\bm{q}_k\in\mathbb{R}^{512},\qquad k=1,\ldots,T_p .
\end{equation}

\vspace{1mm}
\noindent\textbf{Decoder-interface variants.}
Let \(\bm{Z}_t=[\bm{z}_t^1,\ldots,\bm{z}_t^M]\) denote the encoded sensory context grouped by modality.
A decoder layer follows the standard Transformer structure: self-attention among action queries, cross-attention from action queries to sensory tokens, and a feed-forward update.
The self-attention update is shared by both variants,
\begin{equation}
\bar{\bm{h}}_t^l
=
\bm{h}_t^l+\mathrm{SelfAttn}^l(\bm{h}_t^l).
\end{equation}
The flat and group-conditioned decoders differ in the cross-attention update:
\begin{equation}
\begin{array}{@{}r@{\;}c@{\quad}r@{\;}c@{\;}l@{}}
\text{\normalfont\bfseries\ttfamily FlatDec}
&:&
\tilde{\bm{h}}_t^l
&=&
\bar{\bm{h}}_t^l+\mathrm{CrossAttn}^l(\bar{\bm{h}}_t^l,\bm{Z}_t), \\
\text{\normalfont\bfseries\ttfamily GroupDec}
&:&
\tilde{\bm{h}}_t^l
&=&
\bar{\bm{h}}_t^l+\frac{1}{M}\sum_{m=1}^{M}
\mathrm{CrossAttn}_m^l(\bar{\bm{h}}_t^l,\bm{z}_t^m).
\end{array}
\end{equation}
After the cross-attention update, each decoder layer applies the standard Transformer feed-forward update,
\begin{equation}
\bm{h}_t^{l+1}
=
\tilde{\bm{h}}_t^l+\mathrm{FFN}^l(\tilde{\bm{h}}_t^l).
\end{equation}
\textbf{\texttt{FlatDec}} lets all modality tokens compete in one cross-attention normalization.
\textbf{\texttt{GroupDec}} gives each modality group a separate cross-attention path and averages the residual updates, so every group has a direct route into the action queries.
The decoded action features are mapped to joint commands by an MLP head,
\begin{equation}
\hat{\bm{a}}_{t+k-1}
=
f_{\mathrm{act}}(\bm{h}_{t,k}^{L}),\qquad k=1,\ldots,T_p .
\end{equation}

\vspace{1mm}
\noindent\textbf{Architecture-ablation notation.}
The architecture names in the main paper combine the encoder and decoder choices above.
For example, \textbf{\texttt{GEnc+GDec}} uses group-local observation encoding followed by group-conditioned decoder cross-attention, while \textbf{\texttt{FEnc+FDec}} uses flat self-attention and flat cross-attention.
Across these ablations, the tokenizer, action-query initialization, prediction horizon, action head, and training data are held fixed; when needed, hidden or feed-forward widths are adjusted to keep model sizes comparable.

\subsection{Decoder Attention Diagnostics}
\label{app:attention_diagnostics}

To diagnose modality competition at the action interface, we inspect decoder cross-attention from action queries to context tokens in flat-decoder policies.
We group context tokens into \ours{} (\textbf{\texttt{FE}}), \textbf{\texttt{Wrist}}, and \textbf{\texttt{Proprio}} groups, and aggregate attention mass over test-time rollouts on three representative simulation tasks and three evaluation seeds.
We compare \textbf{\texttt{NoEnc+FDec}}, which directly exposes observation tokens to a flat decoder, with \textbf{\texttt{FEnc+FDec}}, an encoded-token flat-decoder baseline that first applies flat self-attention over observation tokens but still uses the same flat decoder readout.

\begin{table}[H]
\centering
\caption{
\textbf{Decoder cross-attention diagnostic for flat decoders.}
We aggregate test-time action-query cross-attention over context tokens and map tokens to \ours{} (\textbf{\texttt{FE}}), \textbf{\texttt{Wrist}}, and \textbf{\texttt{Proprio}} groups. \textit{Prop. win rate} denotes the fraction of action-query samples for which the \textbf{\texttt{Proprio}} group receives the largest group-level attention mass. SR is reported for context from the corresponding main-paper policy-interface evaluation; attention values are computed for train seed \(42\) and averaged over three evaluation seeds. All entries are percentages.
}
\label{tab:flat_decoder_attention}
\scriptsize
\setlength{\tabcolsep}{3pt}
\renewcommand{\arraystretch}{1.05}
\begin{tabular}{@{}llccccc@{}}
\toprule
Task & Policy & SR & \textbf{\texttt{FE}} mass & \textbf{\texttt{Wrist}} mass & \textbf{\texttt{Prop.}} mass & \textbf{\texttt{Prop.}} win rate \\
\midrule
Coin & \textbf{\texttt{NoEnc+FDec}} & \(48.6{\pm}8.3\) & \(51.6{\pm}0.4\) & \(12.9{\pm}0.1\) & \(35.6{\pm}0.4\) & \(18.9{\pm}0.8\) \\
Coin & \textbf{\texttt{FEnc+FDec}} & \(43.6{\pm}8.3\) & \(46.0{\pm}0.1\) & \(10.5{\pm}0.0\) & \(43.4{\pm}0.1\) & \(37.6{\pm}0.4\) \\
Nut Pick & \textbf{\texttt{NoEnc+FDec}} & \(80.7{\pm}11.3\) & \(31.3{\pm}0.2\) & \(9.8{\pm}0.2\) & \(59.0{\pm}0.1\) & \(77.2{\pm}0.4\) \\
Nut Pick & \textbf{\texttt{FEnc+FDec}} & \(83.7{\pm}6.1\) & \(48.8{\pm}0.0\) & \(11.8{\pm}0.0\) & \(39.5{\pm}0.0\) & \(34.1{\pm}0.1\) \\
Peg-in-Hole & \textbf{\texttt{NoEnc+FDec}} & \(26.6{\pm}7.6\) & \(52.0{\pm}0.3\) & \(10.1{\pm}0.1\) & \(38.0{\pm}0.3\) & \(22.8{\pm}0.7\) \\
Peg-in-Hole & \textbf{\texttt{FEnc+FDec}} & \(18.2{\pm}7.5\) & \(42.2{\pm}0.0\) & \(7.8{\pm}0.0\) & \(50.0{\pm}0.1\) & \(61.0{\pm}1.1\) \\
\bottomrule
\end{tabular}
\vspace{-2mm}
\end{table}

Table~\ref{tab:flat_decoder_attention} shows two complementary patterns in the flat-decoder action readout.
First, the single \textbf{\texttt{Proprio}} token receives \(35.6\%\)--\(59.0\%\) of the attention mass, far above the \(16.7\%\) mass it would receive under a token-uniform reference over four \textbf{\texttt{FE}} tokens, one \textbf{\texttt{Wrist}} token, and one \textbf{\texttt{Proprio}} token.
This attention imbalance is consistent with decoder-level modality competition and a proprioceptive shortcut.
Second, the variants with stronger overall SR in the corresponding policy-interface evaluation also show higher \textbf{\texttt{FE}} attention in this diagnostic, suggesting that preserving action-query access to \textbf{\texttt{FE}} observations is important for performance.
On Coin, \textbf{\texttt{NoEnc+FDec}} has higher SR than \textbf{\texttt{FEnc+FDec}} (\(48.6\%\) vs. \(43.6\%\)) and higher \textbf{\texttt{FE}} mass (\(51.6\%\) vs. \(46.0\%\)).
On Nut Pick, \textbf{\texttt{FEnc+FDec}} has higher SR (\(83.7\%\) vs. \(80.7\%\)) and higher \textbf{\texttt{FE}} mass (\(48.8\%\) vs. \(31.3\%\)).
On Peg-in-Hole, \textbf{\texttt{NoEnc+FDec}} again has higher SR (\(26.6\%\) vs. \(18.2\%\)) and higher \textbf{\texttt{FE}} mass (\(52.0\%\) vs. \(42.2\%\)).
Wrist attention remains low across all rows (\(7.8\%\)--\(12.9\%\)), indicating that the main flat-decoder competition is between distributed \textbf{\texttt{FE}} tokens and the \textbf{\texttt{Proprio}} token.

Although attention weights are a diagnostic rather than a causal measure of modality importance, the table suggests that the action readout interface matters: flat decoding can leave \textbf{\texttt{FE}} tokens and the low-dimensional proprioceptive token competing in one cross-attention distribution.
Encoder-side processing can improve token representations, but it does not by itself guarantee that action queries preserve strong \textbf{\texttt{FE}} access during decoding.
This motivates our group-conditioned decoder, which lets action queries read from each modality group separately before combining group-conditioned action features, avoiding a single softmax competition across all modality groups while still fusing \textbf{\texttt{FE}}, \textbf{\texttt{Wrist}}, and \textbf{\texttt{Proprio}} cues for action prediction.

\subsection{Experiment Details}

\subsubsection{Default Policy-Training Settings}
\label{subsec:supp_default_training}
Unless otherwise stated, all policy experiments use the shared training defaults in Table~\ref{tab:default_training_settings}.
The number of training epochs and the prediction/execution horizon \(T_p/T_a\) vary by task and are listed separately in Table~\ref{tab:task_policy_settings}.

\begin{table}[H]
\centering
\caption{Default policy-training settings.}
\label{tab:default_training_settings}
\small
\setlength{\tabcolsep}{5pt}
\renewcommand{\arraystretch}{1.12}
\begin{tabular}{@{}p{0.26\linewidth}p{0.68\linewidth}@{}}
\toprule
Category & Default setting \\
\midrule
Temporal context & \(T_o=1\) observation step; task-specific \(T_p/T_a\) in Table~\ref{tab:task_policy_settings}. \\
Policy backbone & \(4\) encoder layers, \(4\) decoder layers, hidden dimension \(512\), \(16\) attention heads, feed-forward dimension \(2048\), GELU activation, dropout \(0.0\). \\
Modality fusion & Group encoding and group decoding enabled by default, corresponding to \textbf{\texttt{GEnc+GDec}}. \\
Optimizer & AdamW with learning rate \(1.0\times10^{-4}\), \(\beta=(0.95,0.999)\), \(\epsilon=1.0\times10^{-8}\), and weight decay \(1.0\times10^{-6}\). \\
Schedule and batch & Cosine learning-rate schedule, \(500\) warmup steps, batch size \(256\). \\
EMA and evaluation & Exponential moving average enabled with maximum decay \(0.9999\); evaluation uses the final checkpoint. \\
\bottomrule
\end{tabular}
\end{table}

\subsubsection{Task-Specific Training Settings}
\label{subsec:supp_task_settings}
Table~\ref{tab:task_policy_settings} summarizes the task-specific policy settings used to track the experiment configuration.
For each task, all modality ablations and policy-architecture ablations use the same demonstration dataset and the same task-specific train/test configuration split.
Simulation training and testing initial states are sampled from the same randomized ranges.
For real-world tasks, both training and testing randomize the object pose within each listed object, size, angle, or scale condition.
For policy-architecture comparisons, we also size-match the model variants by adjusting hidden or feed-forward widths when needed, so that the comparison reflects the interface design rather than parameter count.

\begin{table}[H]
\centering
\caption{Task-specific policy and dataset settings.}
\label{tab:task_policy_settings}
\scriptsize
\setlength{\tabcolsep}{2pt}
\renewcommand{\arraystretch}{1.13}
\begin{adjustbox}{width=\textwidth,center}
\begin{tabular}{@{}>{\raggedright\arraybackslash}p{0.095\linewidth}
>{\raggedright\arraybackslash}p{0.125\linewidth}
c c c c
>{\raggedright\arraybackslash}p{0.205\linewidth}
>{\raggedright\arraybackslash}p{0.125\linewidth}
>{\raggedright\arraybackslash}p{0.125\linewidth}@{}}
\toprule
Setting & Task & Cams in use & Demos & Epochs & \(T_p/T_a\) & Action space & Train config. & Test config. \\
\midrule
Simulation & Coin standing & \textbf{\texttt{4FE+1W}} & 50 & 500 & \(16/16\) & Abs. hand joints (8) & Randomized coin pose & Same randomized range \\
Simulation & Nut picking & \textbf{\texttt{4FE+1W}} & 50 & 500 & \(16/8\) & Abs. wrist position (3) + abs. hand joints (8) & Randomized nut/support + wrist init & Same randomized range \\
Simulation & Peg-in-Hole pushing & \textbf{\texttt{4FE+1W}} & 200 & 300 & \(16/16\) & Abs. hand joints (8) & Randomized peg/hole offsets & Same randomized range \\
Real-world & Coin standing & \textbf{\texttt{4FE+1W}} & 60 & 500 & \(16/8\) & Abs. hand joints (8) & Rand. pose; coin \(D=10/30/50\) mm & Rand. pose; coin \(D=20/40\) mm \\
Real-world & Chip picking & \textbf{\texttt{4FE+1W}} & 30 & 500 & \(16/8\) & Abs. hand joints (8) + rel. \(z\) (1) & Rand. pose; chip full/half/quarter & Rand. pose; chip full/irregular \\
Real-world & Letter retrieving & \textbf{\texttt{4FE+1W}} & 60 & 500 & \(16/8\) & Abs. hand joints (8) + arm joints (7) & Rand. pose; letter \(A=60^\circ/30^\circ\) & Rand. pose; letter \(A=45^\circ\) \\
Real-world & Syringe manipulation & \textbf{\texttt{6FE+1W}} & 40 & 500 & \(16/8\) & Abs. hand joints (12) + binary arm go-to pose command (1) & Rand. pose; syringe 10/30 mL & Rand. pose; syringe 20 mL \\
\bottomrule
\end{tabular}
\end{adjustbox}
\vspace{1mm}
\begin{minipage}{\textwidth}
\footnotesize
\emph{Note.}
Here, \(4\mathrm{FE}\) denotes four FingerEye camera views, i.e., two FingerEye modules with two cameras each; \(6\mathrm{FE}\) denotes three FingerEye modules with two cameras each. \(1\mathrm{W}\) denotes one wrist RGB camera.
\end{minipage}
\end{table}

\subsubsection{Post-Contact Tactile Map Baseline}
\label{subsec:supp_post_contact_tactile}
This section specifies the \textbf{\texttt{W+T}} simulation baseline used in Q3.
The baseline tests whether post-contact tactile maps can replace continuous \ours observations under the same policy interface.
Unlike \ours, which observes the scene throughout approach, alignment, and contact, \textbf{\texttt{W+T}} builds a tactile proxy from known scene geometry available only in simulation; it is not a PhysX contact-force sensor or a deformable tactile simulation.

For the index and thumb fingertips, we rasterize the local contact surface, treated as a tactile plane, into \(64\times96\) grids.
At each grid point \(g\), the simulator computes the nearest distance \(d_g\) to nearby collision geometry and converts it to a contact-depth response,
\begin{equation}
c_g = \max(0, d_{\mathrm{contact}} - d_g),
\qquad
d_{\mathrm{contact}} = 1~\mathrm{mm}.
\end{equation}
Only geometry within \(1\,\mathrm{mm}\) of the tactile plane produces a nonzero response, so the signal is intentionally silent before near contact.
Each tactile map is encoded by the same frozen RADIO backbone used for RGB observations.
The \textbf{\texttt{W+T}} policy then receives wrist RGB, proprioception, and the two per-finger tactile tokens through the same \textbf{\texttt{GEnc+GDec}} interface as \textbf{\texttt{W+FE}}.

\textbf{\texttt{W+T}} is intentionally a post-contact-only baseline.
After contact, it provides local proximity and contact-geometry cues, but it does not provide the continuous fingertip RGB cues available to \ours during approach, alignment, and edge engagement.
This is consistent with the main-text Q3 result: \textbf{\texttt{W+T}} remains close to wrist-only performance, while \textbf{\texttt{W+FE}} achieves much higher mean success by providing contact-centric observations throughout the full interaction.

\subsubsection{Explicit Wrench-Input Ablation}
\label{subsec:supp_wrench_input}
The main learned policies use RGB observations and proprioception, without explicit plate pose or estimated wrench. To evaluate auxiliary force--torque input, we add a six-dimensional wrench group (\textbf{\texttt{FT}}) in simulation. \textbf{\texttt{W+FT}} achieves \(32.8\%\), \(43.9\%\), and \(23.8\%\) SR on Coin Standing, Nut Picking, and Peg-in-Hole, respectively, yielding \(33.5\%\) mean SR versus \(26.7\%\) for \textbf{\texttt{W}}. Adding this group to \textbf{\texttt{W+FE}} increases mean SR from \(65.9\%\) to \(67.1\%\). Wrench input helps wrist-only policies, but its small incremental gain with \ours suggests substantial redundancy with visually observable deformation cues on these tasks. Force-regulation tasks, such as delicate grasping, may benefit more.

\ifdefined\arxivcombined
\let\suppMaybeEndDocument\relax
\else
{\small
\bibliography{references}

@article{mason2018toward,
  title={Toward robotic manipulation},
  author={Mason, Matthew T},
  journal={Annual Review of Control, Robotics, and Autonomous Systems},
  volume={1},
  number={1},
  pages={1--28},
  year={2018},
  publisher={Annual Reviews}
}

@article{dong2026look,
  title={Look-to-Touch: a vision-enhanced proximity and tactile sensor for distance and geometry perception in robotic manipulation},
  author={Dong, Yueshi and Ren, Jieji and Liu, Zhenle and Peng, Zhanxuan and Yuan, Zihao and Zhang, Ningbin and Gu, Guoying},
  journal={IEEE/ASME Transactions on Mechatronics},
  year={2026},
  publisher={IEEE}
}

@inproceedings{yamaguchi2017implementing,
  title={Implementing tactile behaviors using fingervision},
  author={Yamaguchi, Akihiko and Atkeson, Christopher G},
  booktitle={2017 IEEE-RAS 17th International Conference on Humanoid Robotics (Humanoids)},
  pages={241--248},
  year={2017},
  organization={IEEE}
}

@article{athar2023vistac,
  title={VisTac toward a unified multimodal sensing finger for robotic manipulation},
  author={Athar, Sheeraz and Patel, Gaurav and Xu, Zhengtong and Qiu, Qiang and She, Yu},
  journal={IEEE Sensors Journal},
  volume={23},
  number={20},
  pages={25440--25450},
  year={2023},
  publisher={IEEE}
}

@article{xu2025robopanoptes,
  title={Robopanoptes: The all-seeing robot with whole-body dexterity},
  author={Xu, Xiaomeng and Bauer, Dominik and Song, Shuran},
  journal={arXiv preprint arXiv:2501.05420},
  year={2025}
}

@article{zhu2025shapeforce,
  title={ShapeForce: Low-Cost Soft Robotic Wrist for Contact-Rich Manipulation},
  author={Zhu, Jinxuan and Yan, Zihao and Xiao, Yangyu and Guo, Jingxiang and Tie, Chenrui and Cao, Xinyi and Zheng, Yuhang and Shao, Lin},
  journal={arXiv preprint arXiv:2511.19955},
  year={2025}
}

@inproceedings{ranzinger2024radio,
  title={Am-radio: Agglomerative vision foundation model reduce all domains into one},
  author={Ranzinger, Mike and Heinrich, Greg and Kautz, Jan and Molchanov, Pavlo},
  booktitle={Proceedings of the IEEE/CVF conference on computer vision and pattern recognition},
  pages={12490--12500},
  year={2024}
}

@article{chi2025diffusion,
  title={Diffusion policy: Visuomotor policy learning via action diffusion},
  author={Chi, Cheng and Xu, Zhenjia and Feng, Siyuan and Cousineau, Eric and Du, Yilun and Burchfiel, Benjamin and Tedrake, Russ and Song, Shuran},
  journal={The International Journal of Robotics Research},
  volume={44},
  number={10-11},
  pages={1684--1704},
  year={2025},
  publisher={Sage Publications Sage UK: London, England}
}

@inproceedings{ouyang2020low,
  title={Low-cost fiducial-based 6-axis force-torque sensor},
  author={Ouyang, Rui and Howe, Robert},
  booktitle={2020 IEEE International Conference on Robotics and Automation (ICRA)},
  pages={1653--1659},
  year={2020},
  organization={IEEE}
}

@article{vaswani2017attention,
  title={Attention is all you need},
  author={Vaswani, Ashish and Shazeer, Noam and Parmar, Niki and Uszkoreit, Jakob and Jones, Llion and Gomez, Aidan N and Kaiser, {\L}ukasz and Polosukhin, Illia},
  journal={Advances in neural information processing systems},
  volume={30},
  year={2017}
}

@article{lin20239dtact,
  title={9dtact: A compact vision-based tactile sensor for accurate 3d shape reconstruction and generalizable 6d force estimation},
  author={Lin, Changyi and Zhang, Han and Xu, Jikai and Wu, Lei and Xu, Huazhe},
  journal={IEEE Robotics and Automation Letters},
  year={2023},
  publisher={IEEE}
}

@inproceedings{Ze2024DP3,
	title={3D Diffusion Policy: Generalizable Visuomotor Policy Learning via Simple 3D Representations},
	author={Yanjie Ze and Gu Zhang and Kangning Zhang and Chenyuan Hu and Muhan Wang and Huazhe Xu},
	booktitle={Proceedings of Robotics: Science and Systems (RSS)},
	year={2024}
}

@article{yuan2017gelsight,
  title={Gelsight: High-resolution robot tactile sensors for estimating geometry and force},
  author={Yuan, Wenzhen and Dong, Siyuan and Adelson, Edward H},
  journal={Sensors},
  volume={17},
  number={12},
  pages={2762},
  year={2017},
  publisher={MDPI}
}

@inproceedings{lin2025learning,
  title={Learning visuotactile skills with two multifingered hands},
  author={Lin, Toru and Zhang, Yu and Li, Qiyang and Qi, Haozhi and Yi, Brent and Levine, Sergey and Malik, Jitendra},
  booktitle={2025 IEEE International Conference on Robotics and Automation (ICRA)},
  pages={5637--5643},
  year={2025},
  organization={IEEE}
}

@article{chi2024universal,
  title={Universal manipulation interface: In-the-wild robot teaching without in-the-wild robots},
  author={Chi, Cheng and Xu, Zhenjia and Pan, Chuer and Cousineau, Eric and Burchfiel, Benjamin and Feng, Siyuan and Tedrake, Russ and Song, Shuran},
  journal={arXiv preprint arXiv:2402.10329},
  year={2024}
}

@article{zhao2023act,
  title={Learning fine-grained bimanual manipulation with low-cost hardware},
  author={Zhao, Tony Z and Kumar, Vikash and Levine, Sergey and Finn, Chelsea},
  journal={arXiv preprint arXiv:2304.13705},
  year={2023}
}

@article{act,
  title={Learning fine-grained bimanual manipulation with low-cost hardware},
  author={Zhao, Tony Z and Kumar, Vikash and Levine, Sergey and Finn, Chelsea},
  journal={arXiv preprint arXiv:2304.13705},
  year={2023}
}

@inproceedings{qin2023anyteleop,
  title={AnyTeleop: A General Vision-Based Dexterous Robot Arm-Hand Teleoperation System},
  author={Qin, Yuzhe and Yang, Wei and Huang, Binghao and Van Wyk, Karl and Su, Hao and Wang, Xiaolong and Chao, Yu-Wei and Fox, Dieter},
  booktitle={Robotics: Science and Systems},
  year={2023}
}

@article{li2026simultaneous,
  title={Simultaneous tactile-visual perception for learning multimodal robot manipulation},
  author={Li, Yuyang and Chen, Yinghan and Zhao, Zihang and Li, Puhao and Liu, Tengyu and Huang, Siyuan and Zhu, Yixin},
  journal={IEEE Robotics and Automation Letters},
  year={2026},
  publisher={IEEE}
}

@article{li2025classification,
  title={Classification of Vision-Based Tactile Sensors: A Review},
  author={Li, Haoran and Lin, Yijiong and Lu, Chenghua and Yang, Max and Psomopoulou, Efi and Lepora, Nathan F.},
  journal={IEEE Sensors Journal},
  volume={25},
  number={19},
  pages={35672--35686},
  year={2025},
  publisher={Institute of Electrical and Electronics Engineers (IEEE)},
  doi={10.1109/JSEN.2025.3599236}
}

@inproceedings{xu2025dtactive,
  title={Dtactive: A vision-based tactile sensor with active surface},
  author={Xu, Jikai and Wu, Lei and Lin, Changyi and Zhao, Ding and Xu, Huazhe},
  booktitle={2025 IEEE/RSJ International Conference on Intelligent Robots and Systems (IROS)},
  pages={21664--21670},
  year={2025},
  organization={IEEE}
}

@inproceedings{yamaguchi2016combining,
  title={Combining finger vision and optical tactile sensing: Reducing and handling errors while cutting vegetables},
  author={Yamaguchi, Akihiko and Atkeson, Christopher G},
  booktitle={2016 IEEE-RAS 16th international conference on humanoid robots (humanoids)},
  pages={1045--1051},
  year={2016},
  organization={IEEE}
}

@article{hogan2022finger,
  title={Finger-STS: Combined proximity and tactile sensing for robotic manipulation},
  author={Hogan, Francois R and Tremblay, Jean-Fran{\c{c}}ois and Baghi, Bobak H and Jenkin, Michael and Siddiqi, Kaleem and Dudek, Gregory},
  journal={IEEE Robotics and Automation Letters},
  volume={7},
  number={4},
  pages={10865--10872},
  year={2022},
  publisher={IEEE},
  doi={10.1109/LRA.2022.3191812}
}

@inproceedings{lancaster2022optical,
  title={Optical proximity sensing for pose estimation during in-hand manipulation},
  author={Lancaster, Patrick and Gyawali, Pratik and Mavrogiannis, Christoforos and Srinivasa, Siddhartha S and Smith, Joshua R},
  booktitle={2022 IEEE/RSJ International Conference on Intelligent Robots and Systems (IROS)},
  pages={11818--11825},
  year={2022},
  organization={IEEE}
}

@article{liu2024maniwav,
  title={ManiWAV: Learning Robot Manipulation from In-the-Wild Audio-Visual Data},
  author={Liu, Zeyi and Chi, Cheng and Cousineau, Eric and Kuppuswamy, Naveen and Burchfiel, Benjamin and Song, Shuran},
  journal={arXiv preprint arXiv:2406.19464},
  year={2024}
}

@article{shaw2023leap,
  title={Leap hand: Low-cost, efficient, and anthropomorphic hand for robot learning},
  author={Shaw, Kenneth and Agarwal, Ananye and Pathak, Deepak},
  journal={arXiv preprint arXiv:2309.06440},
  year={2023}
}

@article{ward2018tactip,
  title={The tactip family: Soft optical tactile sensors with 3d-printed biomimetic morphologies},
  author={Ward-Cherrier, Benjamin and Pestell, Nicholas and Cramphorn, Luke and Winstone, Benjamin and Giannaccini, Maria Elena and Rossiter, Jonathan and Lepora, Nathan F},
  journal={Soft robotics},
  volume={5},
  number={2},
  pages={216--227},
  year={2018},
  publisher={Mary Ann Liebert, Inc. 140 Huguenot Street, 3rd Floor New Rochelle, NY 10801 USA}
}

@article{zhao2025polytouch,
  title={PolyTouch: A Robust Multi-Modal Tactile Sensor for Contact-rich Manipulation Using Tactile-Diffusion Policies},
  author={Zhao, Jialiang and Kuppuswamy, Naveen and Feng, Siyuan and Burchfiel, Benjamin and Adelson, Edward},
  journal={arXiv preprint arXiv:2504.19341},
  year={2025}
}

@article{roberge2023stereotac,
  title={StereoTac: A novel visuotactile sensor that combines tactile sensing with 3D vision},
  author={Roberge, Etienne and Fornes, Guillaume and Roberge, Jean-Philippe},
  journal={IEEE Robotics and Automation Letters},
  volume={8},
  number={10},
  pages={6291--6298},
  year={2023},
  publisher={IEEE},
  doi={10.1109/LRA.2023.3304560}
}

@article{ablett2025multimodal,
  title={Multimodal and Force-Matched Imitation Learning with a See-Through Visuotactile Sensor},
  author={Ablett, Trevor and Limoyo, Oliver and Sigal, Adam and Jilani, Affan and Kelly, Jonathan and Siddiqi, Kaleem and Hogan, Francois and Dudek, Gregory},
  journal={IEEE Transactions on Robotics},
  volume={41},
  pages={946--959},
  year={2025},
  doi={10.1109/TRO.2024.3521864}
}

@article{luu2025vision,
  title={Vision-based proximity and tactile sensing for robot arms: Design, perception, and control},
  author={Luu, Quan Khanh and Nguyen, Dinh Quang and Nguyen, Nhan Huu and Dam, Nam Phuong and Ho, Van Anh},
  journal={IEEE Transactions on Robotics},
  volume={41},
  pages={5000--5019},
  year={2025},
  publisher={IEEE},
  doi={10.1109/TRO.2025.3593087}
}

@inproceedings{wang2022spectac,
  title={SpecTac: A visual-tactile dual-modality sensor using UV illumination},
  author={Wang, Qi and Du, Yipai and Wang, Michael Yu},
  booktitle={2022 international conference on robotics and automation (ICRA)},
  pages={10844--10850},
  year={2022},
  organization={IEEE}
}

@article{chang2025spikeatac,
      title={{SpikeATac}: A Multimodal Tactile Finger with Taxelized Dynamic Sensing for Dexterous Manipulation}, 
      author={Chang, Eric T and Ballentine, Peter and He, Zhanpeng and Kim, Do-Gon and Jiang, Kai and Liang, Hua-Hsuan and Palacios, Joaquin and Wang, William and Piacenza, Pedro and Kymissis, Ioannis and Ciocarlie, Matei},
      journal={arXiv preprint arXiv:2510.27048},
      year={2025}
}

@article{yu2022allprinted,
  title={All-printed soft human-machine interface for robotic physicochemical sensing},
  author={Yu, You and Li, Jiahong and Solomon, Samuel A. and Min, Jihong and Tu, Jiaobing and Guo, Wei and Xu, Changhao and Song, Yu and Gao, Wei},
  journal={Science Robotics},
  volume={7},
  number={67},
  pages={eabn0495},
  year={2022},
  doi={10.1126/scirobotics.abn0495}
}

@inproceedings{fishel2012sensing,
  title={Sensing tactile microvibrations with the BioTac—Comparison with human sensitivity},
  author={Fishel, Jeremy A and Loeb, Gerald E},
  booktitle={2012 4th IEEE RAS \& EMBS international conference on biomedical robotics and biomechatronics (BioRob)},
  pages={1122--1127},
  year={2012},
  organization={IEEE}
}

@inproceedings{luu2023soft,
  title={Soft Robotic Link with Controllable Transparency for Vision-based Tactile and Proximity Sensing},
  author={Luu, Quan Khanh and Nguyen, Dinh Quang and Nguyen, Nhan Huu and Ho, Van Anh},
  booktitle={2023 IEEE International Conference on Soft Robotics (RoboSoft)},
  pages={1--6},
  year={2023},
  organization={IEEE},
  doi={10.1109/ROBOSOFT55895.2023.10122059}
}

@article{li2024minitac,
  title={MiniTac: An Ultra-Compact $\text{8 mm}$ Vision-Based Tactile Sensor for Enhanced Palpation in Robot-Assisted Minimally Invasive Surgery},
  author={Li, Wanlin and Zhao, Zihang and Cui, Leiyao and Zhang, Weiyi and Liu, Hangxin and Li, Li-An and Zhu, Yixin},
  journal={IEEE Robotics and Automation Letters},
  volume={9},
  number={12},
  pages={11170--11177},
  year={2024},
  doi={10.1109/LRA.2024.3487516}
}

@article{zhao2025embedding,
  title={Embedding high-resolution touch across robotic hands enables adaptive human-like grasping},
  author={Zhao, Zihang and Li, Wanlin and Li, Yuyang and Liu, Tengyu and Li, Boren and Wang, Meng and Du, Kai and Liu, Hangxin and Zhu, Yixin and Wang, Qining and Althoefer, Kaspar and Zhu, Song-Chun},
  journal={Nature Machine Intelligence},
  volume={7},
  number={6},
  pages={889--900},
  year={2025},
  doi={10.1038/s42256-025-01053-3}
}

@inproceedings{li2022see,
  title={See, Hear, and Feel: Smart Sensory Fusion for Robotic Manipulation},
  author={Li, Hao and Zhang, Yizhi and Zhu, Junzhe and Wang, Shaoxiong and Lee, Michelle A. and Xu, Huazhe and Adelson, Edward and Fei-Fei, Li and Gao, Ruohan and Wu, Jiajun},
  booktitle={Conference on Robot Learning},
  year={2022}
}

@article{feng2024play,
  title={Play to the Score: Stage-Guided Dynamic Multi-Sensory Fusion for Robotic Manipulation},
  author={Feng, Ruoxuan and Hu, Di and Ma, Wenke and Li, Xuelong},
  journal={arXiv preprint arXiv:2408.01366},
  year={2024}
}

@article{higuera2024sparsh,
  title={Sparsh: Self-supervised Touch Representations for Vision-based Tactile Sensing},
  author={Higuera, Carolina and Sharma, Akash and Bodduluri, Chaithanya Krishna and Fan, Taosha and Lancaster, Patrick and Kalakrishnan, Mrinal and Kaess, Michael and Boots, Byron and Lambeta, Mike and Wu, Tingfan and Mukadam, Mustafa},
  journal={arXiv preprint arXiv:2410.24090},
  year={2024}
}

@article{higuera2025tactilebeyondpixels,
  title={Tactile Beyond Pixels: Multisensory Touch Representations for Robot Manipulation},
  author={Higuera, Carolina and Sharma, Akash and Fan, Taosha and Bodduluri, Chaithanya Krishna and Boots, Byron and Kaess, Michael and Lambeta, Mike and Wu, Tingfan and Liu, Zixi and Hogan, Francois Robert and Mukadam, Mustafa},
  journal={arXiv preprint arXiv:2506.14754},
  year={2025}
}

@article{heng2025vitacformer,
  title={ViTacFormer: Learning Cross-Modal Representation for Visuo-Tactile Dexterous Manipulation},
  author={Heng, Liang and Geng, Haoran and Zhang, Kaifeng and Abbeel, Pieter and Malik, Jitendra},
  journal={arXiv preprint arXiv:2506.15953},
  year={2025}
}

@article{liu2025vitamin,
  title={ViTaMIn: Learning Contact-Rich Tasks Through Robot-Free Visuo-Tactile Manipulation Interface},
  author={Liu, Fangchen and Li, Chuanyu and Qin, Yihua and Shaw, Ankit and Xu, Jing and Abbeel, Pieter and Chen, Rui},
  journal={arXiv preprint arXiv:2504.06156},
  year={2025}
}

@article{jiang2025gelfusion,
  title={GelFusion: Enhancing robotic manipulation under visual constraints via visuotactile fusion},
  author={Jiang, Shulong and Zhao, Shiqi and Fan, Yuxuan and Yin, Peng},
  journal={arXiv preprint arXiv:2505.07455},
  year={2025}
}

@article{chen2025multi,
  title={Multi-Modal Manipulation via Multi-Modal Policy Consensus},
  author={Chen, Haonan and Xu, Jiaming and Chen, Hongyu and Hong, Kaiwen and Huang, Binghao and Liu, Chaoqi and Mao, Jiayuan and Li, Yunzhu and Du, Yilun and Driggs-Campbell, Katherine},
  journal={arXiv preprint arXiv:2509.23468},
  year={2025}
}

@article{liu2025factr,
  title={Factr: Force-attending curriculum training for contact-rich policy learning},
  author={Liu, Jason Jingzhou and Li, Yulong and Shaw, Kenneth and Tao, Tony and Salakhutdinov, Ruslan and Pathak, Deepak},
  journal={arXiv preprint arXiv:2502.17432},
  year={2025}
}

@article{lei2026learning,
  title={Learning When to See and When to Feel: Adaptive Vision-Torque Fusion for Contact-Aware Manipulation},
  author={Lei, Jiuzhou and Liu, Chang and She, Yu and Liang, Xiao and Zheng, Minghui},
  journal={arXiv preprint arXiv:2604.01414},
  year={2026}
}

@article{zhu2026touch,
  title={Touch in the wild: Learning fine-grained manipulation with a portable visuo-tactile gripper},
  author={Zhu, Xinyue and Huang, Binghao and Li, Yunzhu},
  journal={Advances in Neural Information Processing Systems},
  volume={38},
  pages={153783--153812},
  year={2026}
}

@article{wang2024dexcap,
  title={Dexcap: Scalable and portable mocap data collection system for dexterous manipulation},
  author={Wang, Chen and Shi, Haochen and Wang, Weizhuo and Zhang, Ruohan and Fei-Fei, Li and Liu, C Karen},
  journal={arXiv preprint arXiv:2403.07788},
  year={2024}
}

@article{lin2025sim,
  title={Sim-to-real reinforcement learning for vision-based dexterous manipulation on humanoids},
  author={Lin, Toru and Sachdev, Kartik and Fan, Linxi and Malik, Jitendra and Zhu, Yuke},
  journal={arXiv preprint arXiv:2502.20396},
  year={2025}
}

@inproceedings{qi2023general,
  title={General in-hand object rotation with vision and touch},
  author={Qi, Haozhi and Yi, Brent and Suresh, Sudharshan and Lambeta, Mike and Ma, Yi and Calandra, Roberto and Malik, Jitendra},
  booktitle={Conference on Robot Learning},
  pages={2549--2564},
  year={2023},
  organization={PMLR}
}

@article{wang2019flexible,
  title={Flexible tactile sensor array for distributed tactile sensing and slip detection in robotic hand grasping},
  author={Wang, Yancheng and Wu, Xin and Mei, Deqing and Zhu, Lingfeng and Chen, Jianing},
  journal={Sensors and Actuators A: Physical},
  volume={297},
  pages={111512},
  year={2019},
  publisher={Elsevier}
}

@article{huang2026flexitac,
  title={FlexiTac: A Low-Cost, Open-Source, Scalable Tactile Sensing Solution for Robotic Systems},
  author={Huang, Binghao and Li, Yunzhu},
  journal={arXiv preprint arXiv:2604.28156},
  year={2026}
}

@article{suresh2024neuralfeels,
  title={NeuralFeels with neural fields: Visuotactile perception for in-hand manipulation},
  author={Suresh, Sudharshan and Qi, Haozhi and Wu, Tingfan and Fan, Taosha and Pineda, Luis and Lambeta, Mike and Malik, Jitendra and Kalakrishnan, Mrinal and Calandra, Roberto and Kaess, Michael and others},
  journal={Science Robotics},
  volume={9},
  number={96},
  pages={eadl0628},
  year={2024},
  publisher={American Association for the Advancement of Science}
}

@article{lin2025lighttact,
  title={LightTact: A Visual-Tactile Fingertip Sensor for Deformation-Independent Contact Sensing},
  author={Lin, Changyi and Huo, Boda and Yu, Mingyang and Ruppel, Emily and Chen, Bingqing and Francis, Jonathan and Zhao, Ding},
  journal={arXiv preprint arXiv:2512.20591},
  year={2025}
}

@article{huang2024high,
  title={A high-sensitivity flexible piezoelectric tactile sensor utilizing an innovative rigid-in-soft structure},
  author={Huang, Xiaodong and Ma, Zeyu and Xia, Wentao and Hao, Luxin and Wu, Yuhao and Lu, Shan and Luo, Yusen and Qin, Liguo and Dong, Guangneng},
  journal={Nano Energy},
  volume={129},
  pages={110019},
  year={2024},
  publisher={Elsevier}
}
}
\def\suppMaybeEndDocument{\end{document}}
\fi
\suppMaybeEndDocument

\end{document}